\documentclass[11pt, logo, copyright, colorlinks=true, allcolors=blue]{nvidiatechreport}

\PassOptionsToPackage{numbers, compress}{natbib}
\usepackage[numbers]{natbib}
\usepackage{xurl}
\usepackage{etoc}
\usepackage{mathtools}
\usepackage{multirow}
\usepackage{capt-of}
\usepackage{placeins}
\usepackage{bm}
\usepackage[most]{tcolorbox}
\usepackage{listings}

\definecolor{remarkgreen}{HTML}{2d6a4f}
\definecolor{claimred}{HTML}{c25e3f}
\definecolor{takeawayblue}{HTML}{1f4e8a}
\definecolor{takeawayfill}{HTML}{F4F7FC}
\definecolor{Green}{rgb}{0,0.4,0.7}
\definecolor{darkgray}{gray}{0.3}
\definecolor{chocolate}{HTML}{D2691E}
\definecolor{green}{HTML}{008000}
\definecolor{mygray}{gray}{0.92}
\definecolor{nvidiagreen}{HTML}{76B900}
\definecolor{dbad}{HTML}{dc2f4a}

\lstdefinestyle{rawexample}{
  basicstyle=\ttfamily\scriptsize,
  breaklines=true,
  columns=fullflexible,
  keepspaces=true,
  showstringspaces=false,
  frame=single,
  framerule=0.35pt,
  rulecolor=\color{darkgray!45},
  backgroundcolor=\color{mygray!45},
  xleftmargin=3pt,
  xrightmargin=3pt,
  aboveskip=4pt,
  belowskip=4pt}
\hypersetup{
    colorlinks=true,
    citecolor=teal,
    linkcolor=nvidiagreen,
    urlcolor=Green,
}

\AtBeginDocument{%
}
\newcommand{\appautoref}[1]{\hyperref[#1]{Appendix~\ref*{#1}}}

\title{Mid-Harness: Scaling Actions Between Model and Harness for Terminal Agents}
\newcommand{\authorblock}{%
Minki Kang$^{1,2*}$,
Ryo Hachiuma$^{1}$,
Shaokun Zhang$^{1}$,
Subhashree Radhakrishnan$^{1}$,
Yonggan Fu$^{1}$,
Jindong Jiang$^{1}$,
Mingjie Liu$^{1}$,
Ehsan Hosseini-Asl$^{1}$,
Yi Dong$^{1}$,
Yu-Chiang Frank Wang$^{1}$,
Byung-Kwan Lee$^{1\dagger}$
}
\author{\authorblock}
\correspondingauthor{$^*$Work done during internship. $^\dagger$Project Lead.}

\renewcommand{\titlefont}{\color{nvidiagreen}\normalfont\bfseries\fontsize{17}{20}\selectfont}

\makeatletter
\renewcommand{\maketitle}{\bgroup\setlength{\parindent}{0pt}
  \begin{adjustwidth}{0pt}{24pt}
  \begin{flushleft}
    {\raggedright \titlefont \@title\par}%
    \vskip11pt
    {\raggedright \normalfont\bfseries\fontsize{9}{13}\selectfont \authorblock\par}%
    \vskip12pt
    {\normalfont\bfseries\fontsize{9}{13}\selectfont $^1$NVIDIA, $^2$KAIST\par}%
    \vskip16pt%
  \end{flushleft}
  \end{adjustwidth}
  \egroup
  \thispagestyle{firststyle}
}
\makeatother

\begin{abstract}
Terminal agents act through stochastic model generations, yet the ability to generate a useful action does not ensure its reliable execution.
A poor command (\emph{e.g.,} wrong package install) can change the environment in ways that hinder subsequent progress, even when the model could generate a better alternative.
We investigate whether allocating test-time compute at the model-harness boundary can improve action reliability and trajectory success, and what makes this allocation effective.
To study these questions, we introduce \textbf{Mid-Harness}, which samples and verifies candidate actions before forwarding one for execution, while keeping the generator and harness unchanged.
With a TMAX-9B generator, more action sampling yields little benefit under weak verification, whereas a capable verifier can exploit useful alternatives from the same generator.
On TerminalBench-Lite, a GPT-5.6 Sol verifier raises Pass@1 from 50.00\% for the base agent to 68.03\% with 8 sampled actions.
When the same TMAX-9B model serves as the verifier, pairwise verification performs best among the evaluated verification mechanisms.
Distilling responses from the stronger verifier into TMAX-9B further improves Pass@1, while leaving the action generator unchanged.
With TMAX-9B on TerminalBench-Lite, combining action and trajectory scaling reaches higher success at lower estimated token cost than generating more trajectories alone.
Mid-Harness also improves performance across additional models, benchmarks, and harnesses.
These findings identify action scaling as a promising target for test-time compute scaling in terminal agents. The project page is available at \href{https://byungkwanlee.github.io/MidHarness-page/}{link}.
\end{abstract}

\begin{document}

\maketitle
\etocdepthtag.toc{main}

\par\addvspace{6pt}
\noindent\begin{minipage}{\linewidth}
  \vspace{-0.14in}
  \centering
  \makebox[\linewidth][c]{%
    \begin{minipage}{\linewidth}
      \centering
      \begin{minipage}[b]{0.50\linewidth}
        \centering
        \begin{minipage}[c][160pt][c]{\linewidth}
          \includegraphics[width=\linewidth,height=160pt,keepaspectratio]{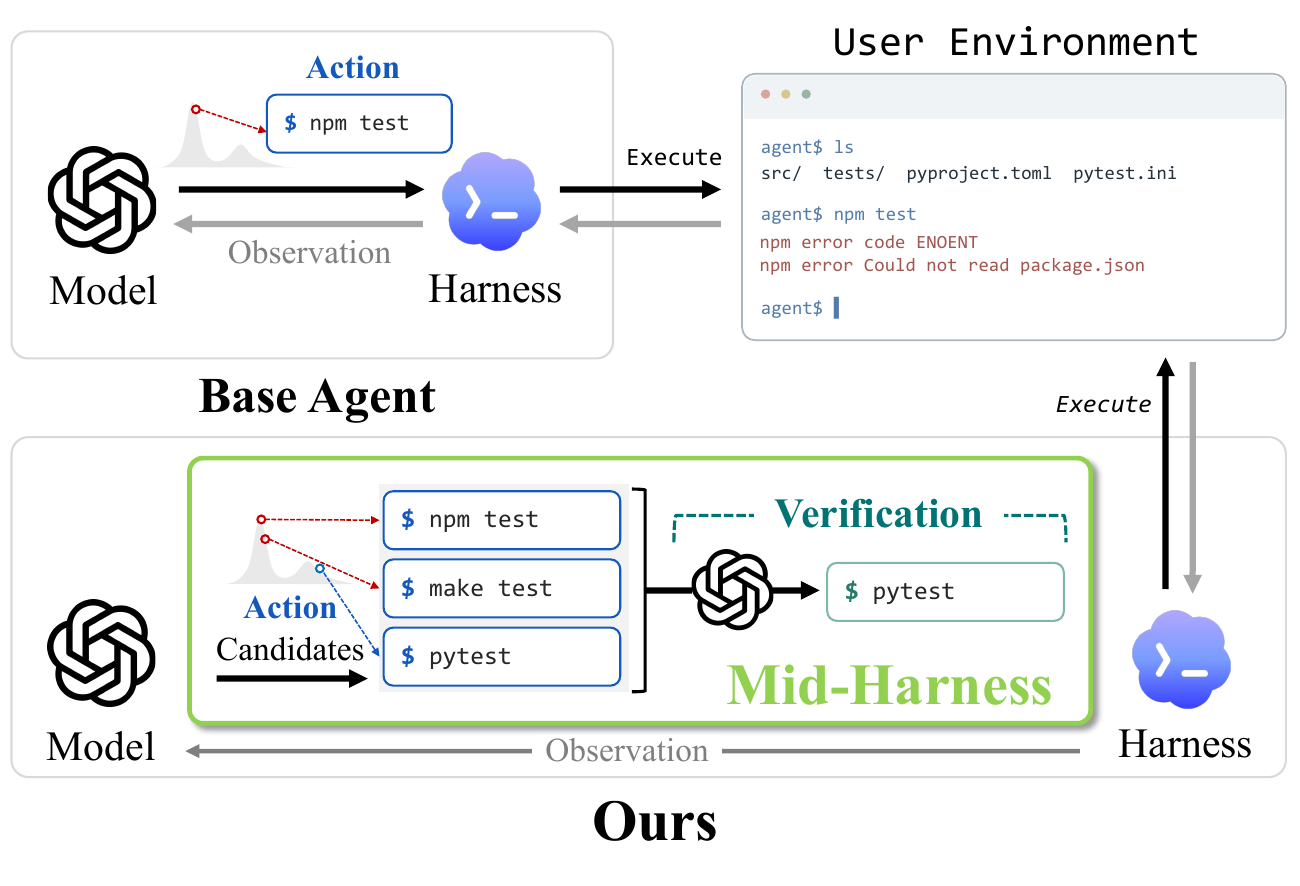}
        \end{minipage}
        \par\smallskip
        \vspace{2pt}
        {\small (a) Concept of Mid-Harness\par}
      \end{minipage}\hfill
      \begin{minipage}[b]{0.50\linewidth}
        \centering
        \begin{minipage}[c][160pt][c]{\linewidth}
          \includegraphics[width=\linewidth,height=160pt,keepaspectratio]{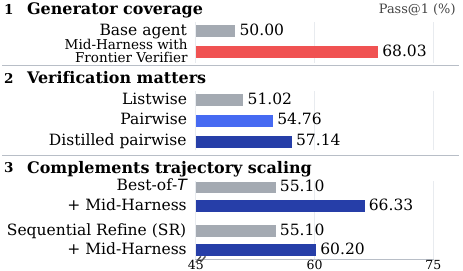}
        \end{minipage}
        \par\smallskip
        \vspace{2pt}
        {\small (b) Summary of empirical findings\par}
      \end{minipage}
    \end{minipage}%
  }
  \captionsetup{hypcap=false}
  \vspace{-0.1in}
  \captionof{figure}{\textbf{(a) Concept figure.} Mid-Harness verifies candidate actions before execution.
\textbf{(b) Effective action verification improves trajectory success and complements trajectory scaling.} TMAX-9B~\citep{tmax} Pass@1 (\%) on TerminalBench-Lite: base versus Mid-Harness with GPT-5.6 Sol verifier (top), results from different verification mechanisms (middle), and combining Mid-Harness with Best-of-$T$ ($T=3$)~\citep{kwok2026llmverifier} and SR ($R=1$)~\citep{SR} (bottom). Mid-Harness uses $N=8$.}
  \label{fig:overview}
\end{minipage}
\par\addvspace{6pt}

\vskip6pt
\section*{Abstract}
{\abscontent}

\section{Introduction}
\label{sec:intro}

Large language models increasingly power terminal agents that perform tasks in software engineering, data science, and scientific discovery~\citep{jimenez2024swebench,tbench2025}.
Yet the ability to generate useful actions does not ensure that an agent completes a task reliably.
The same model and harness may succeed on one run and fail on another because each run commits to a long-horizon sequence of stochastically generated actions.
Each executed command changes the environment on which subsequent decisions depend~\citep{yao2023react,swe-agent,openhands}.
Therefore, a poor command (\textit{e.g.,} wrong code edit, wrong package install) can hinder subsequent progress by changing the environment, even when the model could have generated a better alternative~\citep{RecoveryBench}.
We use \emph{action reliability} to mean consistently generating actions that support task completion, and evaluate its trajectory-level consequences through task success.

Prior work improves reasoning and agent performance by allocating additional test-time compute to sampling, verification, and refinement~\citep{snell2024scaling,cobbe2021verifiers,scalingtesttimeagenticcoding,SR,kwok2026llmverifier}, including verification of candidate actions before execution~\citep{guidedsearch,zhu2025atts}.
These successes motivate scaling compute for action reliability, but leave an incomplete understanding of when and why this scaling improves trajectory success.
In particular, the benefit of generating more candidates depends on the ability to verify them, making it important to study these factors jointly~\citep{brown2024monkeys}.
We therefore conduct a systematic study of action sampling and verification, asking: \emph{when does additional computation before action execution improve trajectory success, and what makes it effective?}

To investigate these questions, we introduce \textbf{Mid-Harness}, a method for studying action-level compute scaling at the model-harness boundary while keeping the action generator and execution harness unchanged (\autoref{fig:overview}(a)).
At each step, Mid-Harness requests several candidate actions from the generator using the same interaction history, applies a verifier, and forwards only the chosen candidate to the harness for execution.
Our main comparisons vary candidate width, verification mechanism, and verifier capability with TMAX-9B~\citep{tmax} and its harness fixed.
We use these comparisons to study two requirements: whether the generator provides useful alternatives, and whether verification can identify them before execution~\citep{mindthegap}.

We first ask whether the generator already produces useful alternatives that could improve trajectory success if reliably identified.
Without ground-truth labels for candidate actions, we probe this opportunity using a strong verifier.
With TMAX-9B as the fixed generator on TerminalBench-Lite~\citep{TBLite}, verification by GPT-5.6 Sol~\citep{GPT5.6} raises Pass@1 from 50.00\% for the base agent to 68.03\% (\autoref{fig:overview}(b)).
This result provides evidence that the generator produces useful alternatives that verification can exploit, improving trajectory success without changing or further post-training the generator.

We next examine how much of this opportunity can be recovered in a \emph{self-verification} setting, where the same model generates and verifies actions.
We find that increasing candidate width yields marginal improvement under weak verification (\autoref{sec:diagnose}).
In this setting, pairwise verification performs best among the evaluated mechanisms, showing that how candidates are compared matters even without changing the verifier model (\autoref{fig:candidate-width}).
Fine-tuning a verifier initialized from the generator on pairwise responses from GPT-5.6 Sol further raises Pass@1 from 54.76\% to 57.14\%, while leaving the action generator unchanged (\autoref{sec:distillation}).
Our analysis shows that distillation increases offline agreement with the frontier verifier, while disagreement over command semantics and execution feasibility persists (\autoref{sec:judgment-analysis}).
Together, these results identify verifying what an action will do in the current environment without executing it as a central challenge in converting candidate diversity into successful trajectories.

Finally, we investigate whether the benefits of action scaling extend to trajectory-level compute scaling methods.
On TerminalBench-Lite, Mid-Harness improves Pass@1 with both parallel scaling using Best-of-$T$ Trajectories with an LLM-as-a-verifier~\citep{kwok2026llmverifier} and sequential scaling using Sequential Refine~\citep{SR} (\autoref{fig:overview}(b), \autoref{sec:composition}).
With TMAX-9B, combining Mid-Harness with one round of SR surpasses Best-of-$T$ at $T=7$ while using less than half its estimated token cost (\autoref{fig:scaling-cost}).
We also find gains across additional models, tasks, and harnesses (\autoref{sec:scale-transfer}).

Our main findings and contributions are:
\vspace{-0.12in}
\begin{itemize}[leftmargin=*,itemsep=1pt]
\item \textbf{Verification governs the benefit of action sampling.}
With TMAX-9B on TerminalBench-Lite, wider sampling yields little benefit under weak verification, while the strong verifier enables substantially more successful trajectories.

\item \textbf{Verification mechanism and training help recover this opportunity.}
Pairwise verification performs best among the evaluated mechanisms, and verifier distillation improves trajectory success without changing the generator.
The offline analysis identifies command semantics and execution feasibility as persistent sources of disagreement with the teacher verifier.

\item \textbf{Mid-Harness enables a systematic study of action scaling in terminal agents.}
By varying sampling and verification at a fixed model-harness boundary, we examine when additional computation improves trajectory success and demonstrate its compatibility with parallel and sequential trajectory scaling.
\end{itemize}

\section{Mid-Harness}
\label{sec:framework}
\label{sec:prelim}

Mid-Harness lets us study how action candidate generation and verification affect trajectory success while keeping the generator and harness fixed.
The pseudocode below shows how Mid-Harness adds candidate sampling and verification to the model-call wrapper, returning one action to the unchanged harness without modifying model weights or serving architecture.

\begin{tcolorbox}[enhanced,colback=white,colframe=black!20,boxrule=0.45pt,
  arc=2pt,left=6pt,right=6pt,top=5pt,bottom=5pt]
\begin{minipage}[t]{0.47\linewidth}
{\small\bfseries Harness loop \normalfont(unchanged)}\par\smallskip
\begin{lstlisting}[language=Python,basicstyle=\ttfamily\fontsize{7.5}{9.2}\selectfont,
  keywordstyle=\color{takeawayblue},columns=fullflexible,keepspaces=true,
  showstringspaces=false,aboveskip=0pt,belowskip=0pt,escapeinside={(*@}{@*)}]
while not done:
    response = await (*@\tikz[remember picture,baseline=(mhcall.base)]\node[inner sep=1pt,rounded corners=1pt,fill=takeawayblue!10,text=takeawayblue] (mhcall) {\strut model.generate};@*)(
        messages)
    actions, done = parse(response)
    obs = await (*@\colorbox{chocolate!10}{\textcolor{chocolate}{env.execute}}@*)(actions)
    messages = update(
        messages, response, obs)
\end{lstlisting}
\end{minipage}\hfill
\begin{minipage}[t]{0.51\linewidth}
{\small\bfseries Model-call wrapper \textcolor{remarkgreen}{(Mid-Harness)}}\par\smallskip
\begingroup
\ttfamily\fontsize{7.5}{9.2}\selectfont
\setlength{\tabcolsep}{2pt}
\renewcommand{\arraystretch}{1.12}
\begin{tabular}{@{}ll@{}}
\multicolumn{2}{@{}l@{}}{\tikz[remember picture,baseline=(mhwrap.base)]\node[inner sep=1pt,rounded corners=1pt,fill=takeawayblue!10,text=takeawayblue] (mhwrap) {\strut async def generate(messages):};}\\
\rowcolor{red!7}\textcolor{claimred}{-} & \textcolor{claimred}{\phantom{xx}return await llm.generate(messages)}\\
\rowcolor{green!7}\textcolor{remarkgreen}{+} & \textcolor{remarkgreen}{\phantom{xx}candidates = await llm.generate(}\\
\rowcolor{green!7}\textcolor{remarkgreen}{+} & \textcolor{remarkgreen}{\phantom{xxxxxx}messages, n=N)}\\
\rowcolor{green!7}\textcolor{remarkgreen}{+} & \textcolor{remarkgreen}{\phantom{xx}return await verify(}\\
\rowcolor{green!7}\textcolor{remarkgreen}{+} & \textcolor{remarkgreen}{\phantom{xxxxxx}llm, messages, candidates)}\\
\end{tabular}
\endgroup
\end{minipage}
\tikz[remember picture,overlay]\draw[->,takeawayblue,line width=0.6pt] ([xshift=5pt]mhcall.east) -- (mhwrap.west);
\par\medskip
{\footnotesize\textbf{Integration pseudocode.} Green additions implement \textcolor{remarkgreen}{Mid-Harness} inside the model-call wrapper.}
\end{tcolorbox}

\subsection{Scaling Actions Within a Trajectory}
\label{sec:mid-harness-loop}

While trajectory-level verification compares completed runs~\citep{kwok2026llmverifier,scalingtesttimeagenticcoding}, Mid-Harness compares actions before execution. Candidates from the same history may differ in purpose or effect, such as diagnosing a failed command versus retrying it.
Comparing candidates against the unresolved task requirement may therefore improve both the next action and the states encountered later, while requiring only one environment instance~\citep{guidedsearch}.

Let $\pi$ be the action-generating model and $h_t=(o_0,a_1,o_1,\ldots,a_{t-1},o_{t-1})$ its interaction history, where $o_0$ contains the task instruction and the later observations contain terminal feedback\footnote{Following ReAct~\citep{yao2023react}, the generator produces reasoning before each sampled action, using \texttt{<think>} tags~\citep{DeepseekR1}. We omit this reasoning from the notation. The verifier receives the task, observed history, and candidate actions, but not the reasoning generated for those candidates.}.
The base agent draws and executes one action from $\pi(\cdot\mid h_t)$.
Mid-Harness samples $N$ candidates conditioned on the same history and identifies one with verifier $\psi$:
\begin{equation}
\mathcal{A}_t=(a_t^1,\ldots,a_t^N),\quad
 a_t^i\sim\pi(\cdot\mid h_t),\qquad
 a_t^\star=\mathrm{Verify}_{\psi}(h_t,\mathcal{A}_t)\in\mathcal{A}_t.
\label{eq:process}
\end{equation}
The verification procedure $\mathrm{Verify}_{\psi}$ can use different mechanisms, but always returns one of the proposed candidates.
The harness executes that action, receives observation $o_t$ from the environment, and updates the history to $h_{t+1}=(h_t,a_t^\star,o_t)$.
The remaining candidates are discarded, and the next set is generated from the updated history.

\subsection{Verification Mechanisms}
\label{sec:verification-protocols}

Given the history $h_t$ and candidate set $\mathcal{A}_t$ in \autoref{eq:process}, a verification mechanism $\mathrm{Verify}_{\psi}$ specifies the verifier calls and how their responses are combined to identify one action for execution.
We compare listwise, pointwise, and pairwise mechanisms to examine how the form of verification affects the use of candidates from a fixed generator.
In each case, the verifier evaluates proposed actions using the task and observed history before their execution.

\paragraph{Listwise verification.}
The verifier receives the entire candidate set in a single prompt and returns a choice, $i^\star=\psi_{\mathrm{list}}(h_t,\mathcal{A}_t)$~\citep{GenSelect}.
\textbf{Pros.} One verifier call exposes all alternatives for direct comparison.
\textbf{Cons.} The verifier must distinguish every candidate and resolve their ranking within the same response.
As the set grows, this joint decision can become difficult even when the set contains a useful action.

\paragraph{Pointwise verification.}
The verifier assigns each candidate a scalar score $q_i=\psi_{\mathrm{point}}(h_t,a_t^i)$ and returns $i^\star\in\arg\max_i q_i$.
This resembles the step-scoring of process reward models~\citep{lightman2023verify,guidedsearch,webshepherd}.
\textbf{Pros.} It decomposes verification into $N$ independent evaluations that can run in parallel.
\textbf{Cons.} Independent scores must place actions with different purposes on a comparable scale, even when an action that appears reasonable in isolation is less useful than another available candidate~\citep{V1}.

\paragraph{Pairwise verification.}
The verifier receives two candidates under the same history and generates a preference with comparative scores~\citep{V1,kwok2026llmverifier}.
For a set of evaluated pairs $\mathcal{C}$, let $J_{ij}=\psi_{\mathrm{pair}}(h_t,a_t^i,a_t^j)$ and $i^\star=\mathrm{Aggregate}(\{J_{ij}:(i,j)\in\mathcal{C}\})$.
The default pairwise verifier ranks candidates by margin-weighted win rates over the evaluated pairs and returns the highest-ranked action.
\textbf{Pros.} It focuses each comparison on a difference between two alternatives and gives the verifier a shared reference.
\textbf{Cons.} Identifying one candidate from the full set requires more model calls than listwise or pointwise verification. For eight candidates, listwise uses one call and pointwise uses eight, while comparing all unordered pairs requires 28 calls.

In the default mechanisms, the verifier $\psi$ generates reasoning followed by scores or a choice, following generative verifiers~\citep{genrm}. Under \emph{self-verification}, the generator $\pi$ and verifier $\psi$ use the same model. Verification details are in \appautoref{app:protocols}.

\section{Experimental Setup}
\label{sec:setup}

We evaluate what makes action scaling effective, whether distillation improves verification, and how this scaling axis composes with compute spent across complete runs.

\paragraph{Tasks and models.}
The main evaluation uses TerminalBench Lite~\citep{TBLite,tbench2025} with a TMAX-9B generator~\citep{tmax} and the Vanillux2 harness. TMAX-9B is trained from Qwen3.5-9B~\citep{qwen3.5} through reinforcement learning on synthetic terminal tasks using Vanillux2.
Model-scale experiments additionally use TMAX-4B and TMAX-27B.
The generator samples at temperature 0.8 with at most 64 steps, a maximum 65,536 token context, and a 16,384 token output limit per step.
Unless otherwise specified, the verifier uses the same language model as the generator.
The frontier verifier in \autoref{sec:coverage} uses GPT-5.6 Sol~\citep{GPT5.6}.

\paragraph{Evaluation metrics.}
A \emph{run} is one execution on a task, and its \emph{trajectory} is the resulting sequence of actions and observations.
We evaluate three runs per task and report \textbf{Pass@1} as the average exact-success rate across those runs.
\textbf{Pass@3} measures the fraction of tasks solved by at least one of the three runs.
For Best-of-$T$, Pass@1 evaluates the returned output for each task.
More evaluation details are in \appautoref{app:evaluation}.

\paragraph{Baselines and compute-scaling methods.}
The \emph{base agent} executes one sampled action per step in a single run, without additional action verification or trajectory scaling.
We organize additional compute at two levels: before action execution and across complete runs.

The action-level axis applies zero-shot or distilled Mid-Harness inside a run.
Zero-shot Mid-Harness uses the same generator model as the verifier without any fine-tuning, while distilled Mid-Harness uses a verifier with the model fine-tuned on pairwise comparisons from GPT-5.6 Sol as described in \autoref{sec:distillation}.
At the trajectory level, \emph{Best-of-$T$ Trajectories (Best-of-$T$)} uses a probabilistic pivot tournament to choose one output from $T$ completed trajectories~\citep{kwok2026llmverifier}.
The corresponding TMAX model serves as the trajectory verifier at each model scale.
The sequential trajectory-level method \emph{Sequential Refine (SR)}~\citep{SR, scalingtesttimeagenticcoding} performs $R$ refinement rounds, each summarizing the previous run to guide a fresh run.
The corresponding TMAX model produces the trajectory summary.
Further baseline details are provided in \appautoref{app:evaluation}.

\paragraph{Inference cost.}
\emph{Parallelized output tokens (POT)} approximate decoding latency of both generator and verifier under idealized parallel execution.
\emph{Verifier total output tokens} instead sum the outputs of all verifier calls without parallelization.
\autoref{fig:cost-overview} presents these two measures. For \autoref{fig:scaling-cost}, we apply \emph{reference token prices} to generator and verifier tokens as a proxy.
Details are in~\appautoref{app:compute}.

\section{When Does Action Scaling Work?}
\label{sec:diagnose}
\label{sec:reliable}

We examine when sampling and verification improve trajectory success, their costs, and the effect of verifier distillation, with a fixed TMAX-9B generator.

\begin{figure}[t]
\centering
\includegraphics[width=\linewidth]{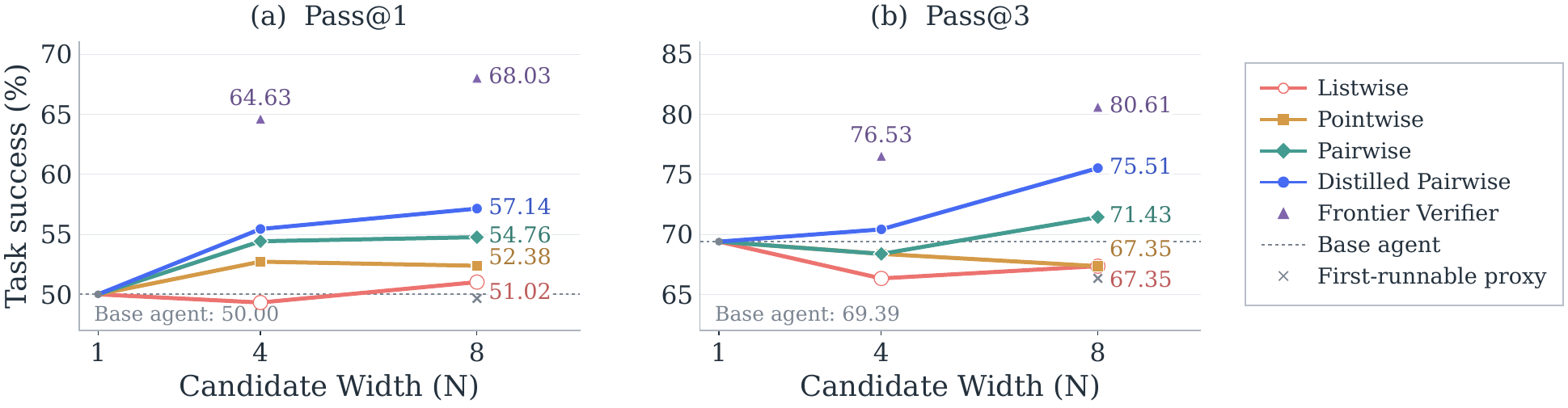}
\vspace{-0.2in}
\caption{\textbf{Verification mechanism determines the return from more action candidates.}
Pass@1 and Pass@3 on TerminalBench-Lite with TMAX-9B. $N=1$ denotes the base agent without action scaling. For frontier verifier, we use the listwise verification due to its cost.}
\vspace{-0.12in}
\label{fig:candidate-width}
\end{figure}

\subsection{Candidate Coverage}
\label{sec:coverage}

\paragraph{A frontier verifier reveals exploitable coverage in the sampled actions.}
We test whether sampled actions contain useful alternatives by pairing the fixed TMAX-9B generator with GPT-5.6 Sol~\citep{GPT5.6}.
Candidate coverage concerns whether sampled actions include useful alternatives. Without action-level ground truth, we probe it indirectly through trajectory success under a strong verifier. As shown in~\autoref{fig:candidate-width}, this configuration reaches 64.63\% Pass@1 at $N=4$ and 68.03\% at $N=8$, showing that the generator's action candidates support substantially more reliable trajectories.
We next ask how much of this opportunity self-verification can recover.

\subsection{Candidate Width and Verification}
\label{sec:candidate_width}

\paragraph{More candidates do not compensate for weak verification.}
Under zero-shot listwise verification, doubling the width from $N=4$ to $N=8$ changes Pass@1 only from 49.32\% to 51.02\% and Pass@3 from 66.33\% to 67.35\% (\autoref{fig:candidate-width}).
The frontier verifier uses the same listwise mechanism but achieves substantially higher success (\autoref{sec:coverage}).
This contrast suggests that the weaker zero-shot verifier struggles to distinguish useful actions when comparing the full candidate set at once, so additional candidates alone offer little benefit.

\paragraph{Pairwise verification performs best among the evaluated mechanisms.}
With the generator and $N=8$ fixed, pointwise improves Pass@1 only slightly over listwise and leaves Pass@3 unchanged, while pairwise reaches 54.76\% Pass@1 and 71.43\% Pass@3 (\autoref{fig:candidate-width}).

\subsection{Verifier Distillation}
\label{sec:distillation}

\paragraph{Distillation setup.}
The frontier verifier result reveals a substantial gap between GPT-5.6 Sol and the generator model (TMAX-9B) as a verifier.
We test whether supervised distillation can transfer part of this capability without serving the frontier model at inference time~\citep{skd}.
Using 117k frontier-verifier pairwise responses collected from 732 trajectories across 244 difficult TMAX-15k tasks~\citep{tmax}, we use LoRA~\citep{hu2021lora} to train a verifier to generate the GPT-5.6 Sol teacher's reasoning, scores, and preference label.
Fine-tuned LoRA is only activated for the verifier, not the generator.
Training data and optimization details are provided in \appautoref{app:training}.

\paragraph{Distillation narrows the verifier-quality gap.}
As shown in~\autoref{fig:candidate-width}, distillation at $N=8$ raises Pass@1 from 54.76\% to 57.14\% and Pass@3 from 71.43\% to 75.51\% for pairwise verification.
The improvement also appears at $N=4$, where Pass@1 rises from 54.42\% to 55.44\% and Pass@3 rises from 68.37\% to 70.41\%.
Even the strongest evaluated zero-shot mechanism leaves room for improvement: verifier distillation raises trajectory success without changing the generator.
We next examine which pairwise preferences distillation improves and which remain difficult (\autoref{sec:judgment-analysis}).

\begin{figure}[t]
\centering
\includegraphics[width=\linewidth]{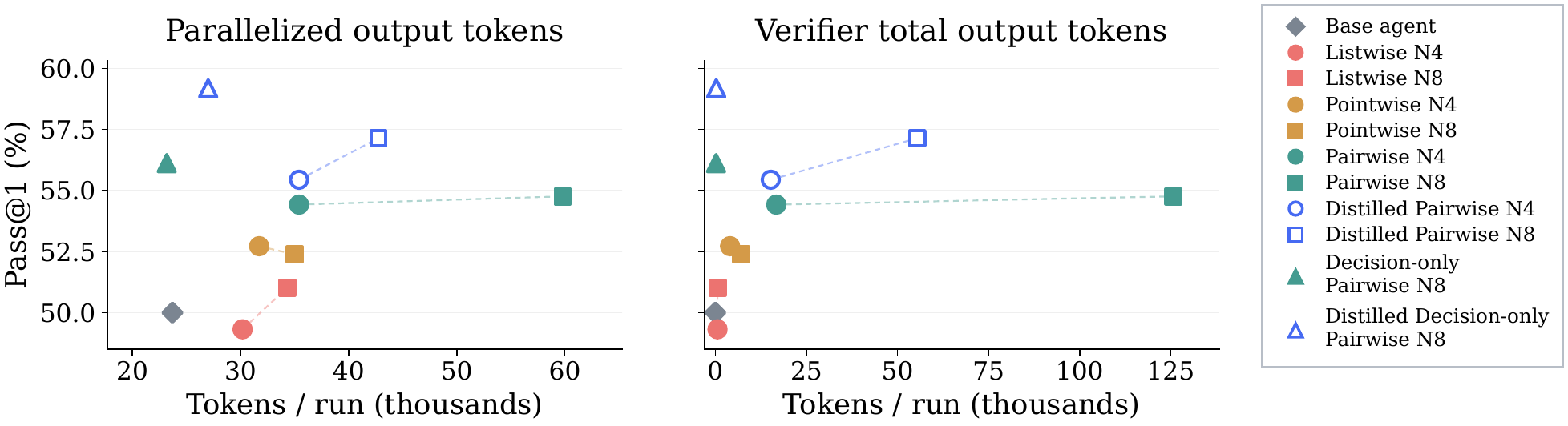}
\vspace{-0.2in}
\caption{\textbf{Verification mechanisms trade off trajectory success against estimated decoding latency and verifier output-token cost.}
Pass@1 versus parallelized output tokens (left) and total verifier output tokens (right) per run on TerminalBench-Lite with TMAX-9B. Dashed lines connect $N=4$ and $N=8$ within each mechanism. Triangles show decision-only pairwise verification at $N=8$. Cost estimates follow \appautoref{app:compute}.}
\vspace{-0.15in}
\label{fig:cost-overview}
\end{figure}

\subsection{Cost for Action Scaling and Verification}
\label{sec:costforscaling}

\paragraph{Pairwise verification adds substantial decoding cost.}
\autoref{fig:cost-overview} presents parallelized output tokens (POT), an overall idealized decoding latency proxy, and total verifier output tokens.
On the left, widening from $N=4$ to $N=8$ increases generator POT since the longest of more sampled generations tends to be longer.
Pairwise verification also adds more verifier decoding cost. At $N=8$, zero-shot pairwise responses add an estimated $26.6k$ POT, versus at most $1.3k$ for listwise and pointwise.
On the right, total verifier output at $N=8$ is $0.6k$ for listwise, $7.0k$ for pointwise, and $125.7k$ for zero-shot pairwise verification.

\paragraph{Effective verification may not require reasoning.}
Does effective action verification require explicit reasoning generation~\citep{genrm}?
Following direct prediction in discriminative reward models~\citep{lightman2023verify,mathsheperd}, we evaluate pairwise verifiers that emit only an \texttt{A}/\texttt{B} preference (Decision-only setting), using the zero-shot model or a model distilled for one-token responses.
At $N=8$, both TMAX-9B decision-only verifiers improve Pass@1 over their reasoning counterparts while lowering reference-priced token cost by 20.9\% and 24.1\% for zero-shot and distilled verification, respectively (\autoref{tab:decision-only-cost}).
At 4B and 27B, the evaluated decision-only variants have lower estimated token cost but also lower Pass@1 than their reasoning counterparts (\autoref{tab:decision-only-cost}).
Additional Pass@3 and $N=4$ results appear in \appautoref{app:plot-results}.

\FloatBarrier

\section{Analysis: What Still Limits Verification?}
\label{sec:judgment-analysis}

We analyze what distillation transfers from the GPT-5.6 Sol teacher using stored TMAX-9B trajectories from 21 held-out TMAX-15K tasks~\citep{tmax}, with $N=8$.
We use this \emph{offline benchmark} to compare pairwise verification outputs of different models under the same state.
These diagnostics measure teacher agreement on verification, not action correctness or trajectory success.

\emph{Pairwise agreement} is the fraction of comparisons with matching verifier and teacher A/B/TIE preferences.
For \emph{verification agreement}, we count each candidate's wins in the pairwise comparisons separately for both models, then measure the fraction of states where they identify the same top-ranked candidate.
\appautoref{app:matched-diagnostics} provides calculation details.

\noindent\begin{minipage}{\linewidth}
\subsection{What Distillation Improves}
\label{sec:distillation-recovery}

\noindent\begin{minipage}[t]{0.60\linewidth}
\vspace{0pt}
\paragraph{Distillation transfers the frontier verifier's behavior.}
On the offline benchmark, distillation reduces the score MAE of pairwise comparisons from 2.59 to 1.05 and raises pairwise agreement from 59.01\% to 74.58\% (\autoref{tab:agreement-summary}).
Verification agreement rises from 38.52\% to 57.79\%, showing that the candidate identified by offline verification more often matches the frontier verifier's candidate after distillation.
\end{minipage}\hfill
\begin{minipage}[t]{0.39\linewidth}
\vspace{0pt}
\centering
\captionsetup{hypcap=false}
\captionof{table}{\textbf{Distillation improves frontier-verifier agreement.}}
\vspace{-0.15in}
\label{tab:agreement-summary}
\small
\setlength{\tabcolsep}{1.5pt}
\begin{tabular}{@{}lrrr@{}}
\toprule
\textbf{Verifier} & \shortstack{\textbf{Score}\\\textbf{MAE}} & \shortstack{\textbf{Pairwise}\\\textbf{(\%)}} & \shortstack{\textbf{Verification}\\\textbf{(\%)}}\\
\midrule
Zero-shot & 2.59 & 59.01 & 38.52\\
Distilled & 1.05 & 74.58 & 57.79\\
\bottomrule
\end{tabular}

\end{minipage}
\end{minipage}

\begin{figure}[!t]
\centering
\includegraphics[width=0.95\linewidth]{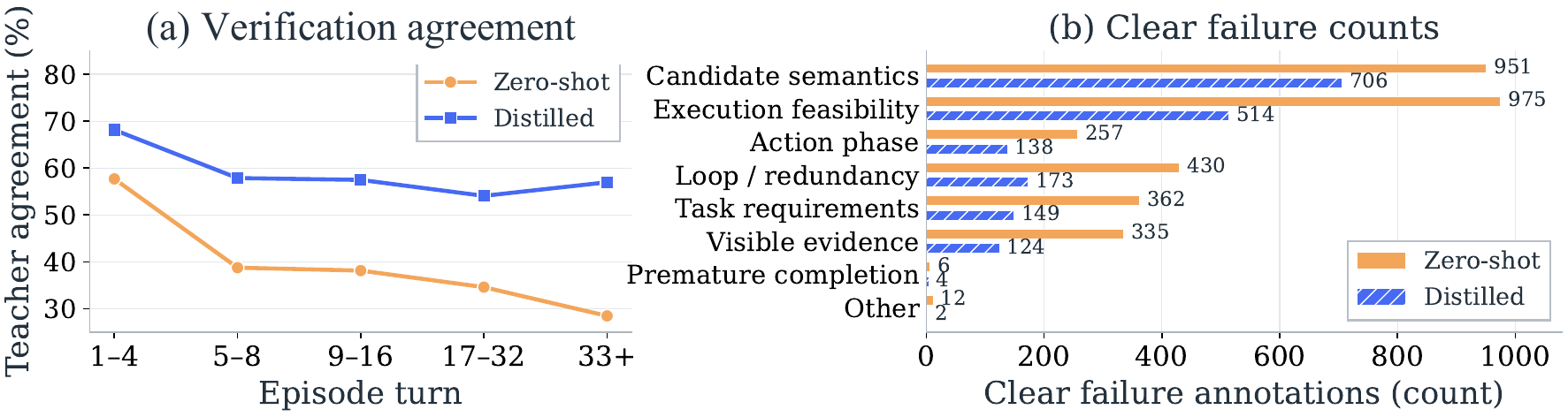}
\vspace{-0.12in}
\caption{\textbf{Distillation improves teacher agreement, but command reasoning remains difficult.}
(a) Offline verification agreement with the frontier verifier by episode turn, using 1,355 states with valid comparisons for both models (83.4\% of states).
(b) Teacher disagreements judged clear verifier failures by GPT-5.6 Terra, grouped by category.}
\vspace{-0.15in}
\label{fig:verifier-diagnostics}
\end{figure}

\subsection{Remaining Verification Challenges}
\label{sec:distillation-remaining}

\paragraph{Agreement improves throughout the trajectory but remains lower in later states.}
Verification agreement improves in every turn bin (\autoref{fig:verifier-diagnostics}(a)).
It reaches 68.13\% at turns 1-4 and 54.07\% at turns 17-32, showing that the gains persist beyond the opening decisions while substantial disagreement remains later in the trajectory.

\paragraph{Remaining disagreements center on command semantics and execution feasibility.}
We use GPT-5.6 Terra~\citep{GPT5.6} to review teacher disagreements and classify those it judges clear verifier failures using a predefined taxonomy.
As in \autoref{fig:verifier-diagnostics}(b), the number of such cases falls from 3,328 for zero-shot verification to 1,810 after distillation, with fewer cases in every category.
Candidate semantics and execution feasibility account for 67.4\% of these distilled-verifier cases.
These categories concern judging command effects in the current environment (\appautoref{app:residual}), illustrated by the executed traces in \appautoref{app:qualitative-examples}.

\section{Composition and Transfer of Action Scaling}
\label{sec:composition-transfer}

We now test whether action verification complements trajectory scaling and transfers across models, harnesses, and tasks.
Best-of-$T$ and SR require fresh environment runs, which can be difficult outside benchmarks without reliable state serialization~\citep{criu,guidedsearch}.
We test whether Mid-Harness improves both methods without increasing their number of environment runs.

\subsection{Composition with Trajectory Scaling}
\label{sec:composition}

\autoref{tab:scaling-axes} compares action and trajectory scaling across TMAX-4B, 9B, and 27B, with task-level confidence intervals in \appautoref{app:uncertainty}.

\paragraph{Mid-Harness improves the trajectory for parallel scaling.}
With TMAX-9B as both generator and trajectory verifier, Best-of-$T$ with $T=3$ reaches 55.10\% Pass@1.
Using zero-shot or distilled Mid-Harness to generate those runs raises Pass@1 to 61.22\% and 66.33\%, respectively (\autoref{tab:scaling-axes}).
This is an 11.23 pp gain with the same three environment executions compared to Best-of-$T$ alone.

\paragraph{Mid-Harness also improves sequentially refined trajectories.}
Using distilled Mid-Harness for both the source trajectories and their refinements raises Pass@1 from 55.10\% with baseline SR to 60.20\%, while Pass@3 rises from 71.43\% to 75.51\%.
Action verification therefore remains useful when a trajectory is conditioned on experience from an earlier run.

\begin{figure}[!t]
\centering
\begin{minipage}{\linewidth}
\captionsetup{type=table}
\centering\small
\captionsetup{width=1.00\textwidth}
\definecolor{axischeck}{HTML}{087F5B}
\caption{\textbf{Composing action and trajectory scaling.}
Pass@1 and Pass@3 (\%) on TerminalBench-Lite with fixed TMAX generators. Mid-Harness uses pairwise verification ($N=8$), Best-of-$T$ uses $T=3$ trajectories, and SR uses $R=1$ round. \textcolor{dbad}{$-\Delta$} and \textcolor{axischeck}{$+\Delta$} show changes from the base agent without any scaling. Checkmarks \textcolor{axischeck}{\ensuremath{\bm{\checkmark}}} indicate enabled components. \# Env. counts environment executions per returned output. Best-of-$T$ returns one output, so Pass@3 is undefined.}
\vspace{-0.1in}
\label{tab:scaling-axes}
\setlength{\tabcolsep}{3.5pt}
\definecolor{axischeckbg}{HTML}{E6F4EE}
\newcommand{\axisyes}{\cellcolor{axischeckbg}\textcolor{axischeck}{\ensuremath{\bm{\checkmark}}}}
\newcommand{\axisno}{\textcolor{darkgray}{\ensuremath{\times}}}
\newcommand{\perfpos}[1]{\,\raisebox{-0.45ex}{\scriptsize\textcolor{axischeck}{\ensuremath{+#1}}}}
\newcommand{\perfneg}[1]{\,\raisebox{-0.45ex}{\scriptsize\textcolor{dbad}{\ensuremath{-#1}}}}
\newcommand{\perfzero}{\,\raisebox{-0.45ex}{\scriptsize\textcolor{darkgray}{\ensuremath{+0.00}}}}
\resizebox{1.00\textwidth}{!}{%
\begin{tabular}{@{}ccccrrrrrrc@{}}
\toprule
\multicolumn{2}{c}{\textbf{Trajectory-level}} &
\multicolumn{2}{c}{\textbf{Action-level}} &
\multicolumn{2}{c}{\textbf{4B}} &
\multicolumn{2}{c}{\textbf{9B}} &
\multicolumn{2}{c}{\textbf{27B}} &
\multirow{2}{*}{\textbf{\# Env.}}\\
\cmidrule(lr){1-2}\cmidrule(lr){3-4}\cmidrule(lr){5-6}\cmidrule(lr){7-8}\cmidrule(lr){9-10}
\multirow{2}{*}{\textbf{Best-of-$T$}} &
\multirow{2}{*}{\textbf{SR}} &
\multicolumn{2}{c}{\textbf{Mid-Harness}} &
\multirow{2}{*}{\textbf{Pass@1}} & \multirow{2}{*}{\textbf{Pass@3}} &
\multirow{2}{*}{\textbf{Pass@1}} & \multirow{2}{*}{\textbf{Pass@3}} &
\multirow{2}{*}{\textbf{Pass@1}} & \multirow{2}{*}{\textbf{Pass@3}} & \\
& & \textbf{zero-shot} & \textbf{distilled} & & & & & & & \\
\midrule
\axisno & \axisno & \axisno & \axisno & 38.78 & 57.14 & 50.00 & 69.39 & 71.09 & 82.65 & 1\\
\axisyes & \axisno & \axisno & \axisno & 34.69\perfneg{4.09} & -- & 55.10\perfpos{5.10} & -- & 73.47\perfpos{2.38} & -- & 3\\
\axisno & \axisyes & \axisno & \axisno & 41.50\perfpos{2.72} & 57.14\perfzero & 55.10\perfpos{5.10} & 71.43\perfpos{2.04} & 72.79\perfpos{1.70} & 84.69\perfpos{2.04} & 2\\
\midrule
\axisno & \axisno & \axisyes & \axisno & 41.50\perfpos{2.72} & 57.14\perfzero & 54.76\perfpos{4.76} & 71.43\perfpos{2.04} & 73.13\perfpos{2.04} & 84.69\perfpos{2.04} & 1\\
\axisyes & \axisno & \axisyes & \axisno & 39.80\perfpos{1.02} & -- & 61.22\perfpos{11.22} & -- & 77.55\perfpos{6.46} & -- & 3\\
\axisno & \axisyes & \axisyes & \axisno & 44.22\perfpos{5.44} & 59.18\perfpos{2.04} & 56.80\perfpos{6.80} & 73.47\perfpos{4.08} & 74.15\perfpos{3.06} & 82.65\perfzero & 2\\
\midrule
\axisno & \axisno & \axisno & \axisyes & 43.88\perfpos{5.10} & 58.16\perfpos{1.02} & 57.14\perfpos{7.14} & \textbf{75.51}\perfpos{6.12} & 76.19\perfpos{5.10} & \textbf{86.73}\perfpos{4.08} & 1\\
\axisyes & \axisno & \axisno & \axisyes & \textbf{46.94}\perfpos{8.16} & -- & \textbf{66.33}\perfpos{16.33} & -- & \textbf{80.61}\perfpos{9.52} & -- & 3\\
\axisno & \axisyes & \axisno & \axisyes & 46.60\perfpos{7.82} & \textbf{62.24}\perfpos{5.10} & 60.20\perfpos{10.20} & \textbf{75.51}\perfpos{6.12} & 75.85\perfpos{4.76} & 85.71\perfpos{3.06} & 2\\
\bottomrule
\end{tabular}}
\end{minipage}

\par\smallskip
\input{figures/fig_scaling_cost}
\end{figure}

\paragraph{Combining scaling axes improves the cost-success trade-off.}
\autoref{fig:scaling-cost} reveals three patterns with TMAX-9B.
First, SR uses relatively little token compute, but Pass@1 plateaus at 55.10\%, 56.46\%, and 55.78\% over $R=1,2,3$.
Second, distilled Mid-Harness at $N=8$ matches Best-of-$T$ at $T=5$ (57.14\%) at about one-third of its reference-priced token cost.
Third, combining it with SR or Best-of-$T$ at $T=3$ reaches 60.20\% or 66.33\%, respectively, both exceeding Best-of-$T$ at $T=7$ (59.18\%) at lower reference-priced token cost.
Thus, combining action and trajectory scaling can achieve higher success at lower estimated token cost than increasing trajectory count alone.
Using decision-only verifiers further reduces the total reference-priced token cost of these compositions by 22-24\% (\appautoref{app:decision-only-cost}).

\subsection{Transfer Across Models, Benchmarks, Harnesses}
\label{sec:scale-transfer}

\paragraph{Action scaling improves generators from 4B to 27B.}
At 4B, zero-shot Mid-Harness raises Pass@1 from 38.78\% to 41.50\%, and distillation raises it to 43.88\%.
At 27B, the corresponding progression is 71.09\%, 73.13\%, and 76.19\%.
Each distilled verifier is initialized from the generator backbone used at each scale, while the pairwise mechanism remains unchanged.

\begin{table}[t]
\centering\small
\captionsetup{width=1.00\textwidth}
\definecolor{transfergreen}{HTML}{087F5B}
\caption{\textbf{Mid-Harness transfers across benchmarks, models, and harnesses.}
Pass@1 / Pass@3 over three runs per task. SWE-bench-Verified uses its Mini subset (50 tasks), and FeatureBench-Mini uses 23 CPU tasks. Both Mid-Harness variants use pairwise verification with $N=8$. Green subscripts \textcolor{transfergreen}{$+\Delta$} show differences against the base agent.}
\vspace{-0.11in}
\label{tab:transfer}
\newcommand{\transfergain}[1]{\,\raisebox{-0.45ex}{\scriptsize\textcolor{transfergreen}{\ensuremath{+#1}}}}
\resizebox{1.00\textwidth}{!}{%
\begin{tabular}{lllccc}
\toprule
\multirow{2}{*}{\textbf{Benchmark}} & \multirow{2}{*}{\textbf{Model}} & \multirow{2}{*}{\textbf{Harness}} & \multirow{2}{*}{\textbf{Base agent}} & \multicolumn{2}{c}{\textbf{Mid-Harness}}\\
\cmidrule(lr){5-6}
& & & & \textbf{Zero-shot} & \textbf{Distilled}\\
\midrule
TerminalBench-Lite & Qwen3.5-9B & Terminus-2 & 40.48 / 60.20 & \textbf{42.52}\transfergain{2.04} / \textbf{61.22}\transfergain{1.02} & --\\
TerminalBench-Lite & Nemotron3.5 Lightning & Terminus-2 & 41.16 / 55.10 & \textbf{43.20}\transfergain{2.04} / \textbf{58.16}\transfergain{3.06} & --\\
\addlinespace[1.5pt]
Terminal-Bench 2.1 & TMAX-9B & Vanillux2 & 21.72 / 25.84 & \textbf{27.34}\transfergain{5.62} / \textbf{35.96}\transfergain{10.12} & 26.59\transfergain{4.87} / \textbf{35.96}\transfergain{10.12}\\
Terminal-Bench 2.1 & Nemotron3 Ultra & Terminus-2 & 50.94 / 65.17 & \textbf{56.18}\transfergain{5.24} / \textbf{66.29}\transfergain{1.12} & --\\
\addlinespace[1.5pt]
SWE-bench-Verified & TMAX-9B & Vanillux2 & 46.67 / 54.00 & 48.00\transfergain{1.33} / 58.00\transfergain{4.00} & \textbf{48.67}\transfergain{2.00} / \textbf{62.00}\transfergain{8.00}\\
\addlinespace[1.5pt]
FeatureBench-Mini & TMAX-9B & Vanillux2 & 1.45 / 4.35 & 5.80\transfergain{4.35} / \textbf{17.39}\transfergain{13.04} & \textbf{7.25}\transfergain{5.80} / 13.04\transfergain{8.69}\\
FeatureBench-Mini & TMAX-27B & Vanillux2 & 17.39 / 26.09 & 17.39 / 34.78\transfergain{8.70} & \textbf{23.19}\transfergain{5.80} / \textbf{39.13}\transfergain{13.04}\\
\bottomrule
\end{tabular}%
}
\vspace{-0.2in}
\end{table}

\paragraph{Action scaling improves success on challenging benchmarks.}
On Terminal-Bench 2.1~\citep{tbench2025}, zero-shot verification raises TMAX-9B Pass@1 from 21.72\% to 27.34\% (\autoref{tab:transfer}).
FeatureBench-Mini~\citep{featurebench} is particularly challenging for TMAX-9B: the base agent succeeds in only 1.45\% of runs, yet zero-shot and distilled verification raise Pass@1 to 5.80\% and 7.25\%, respectively.
The gains extend to TMAX-27B, where distilled verification raises FeatureBench-Mini Pass@1 from 17.39\% to 23.19\% and Pass@3 from 26.09\% to 39.13\%.
Thus, action verification improves success even on tasks where the base agent rarely succeeds.
Both variants also improve TMAX-9B on SWE-bench-Verified~\citep{jimenez2024swebench}.
For TMAX-9B, distillation improves FeatureBench-Mini Pass@1 but lowers Pass@3 relative to zero-shot verification, and does not improve over zero-shot verification on Terminal-Bench 2.1.

\paragraph{The gains appear with other models and a different harness.}
Zero-shot Mid-Harness improves both metrics for Nemotron3.5 Lightning (30B)~\citep{nemo-lightning} and Nemotron3 Ultra (550B)~\citep{nemo-ultra}, extending beyond the TMAX family and model sizes.
It also improves Qwen3.5-9B without additional terminal-specific RL, using Terminus-2~\citep{tbench2025}.
Across all seven settings, zero-shot Mid-Harness improves Pass@3 and matches or improves Pass@1.
Evaluation details and task-group analyses are in \appautoref{app:transfer-evaluation} and \appautoref{app:stratified-scaling}.

\section{Related Work}
\label{sec:related}

\paragraph{Models and harnesses in terminal agents.}
Prior work develops agent interfaces~\citep{swe-agent, openhands}, context management~\citep{primeagent}, and automated harness optimization~\citep{meta-harness}.
Model-side work scales supervised and reinforcement learning with synthetic tasks~\citep{nemotronterminal,openthoughts,endlessterminals,tmax,frognano}.
Mid-Harness studies action scaling with the model and harness fixed.

\paragraph{Test-time scaling and process reward models.}
Test-time compute scaling improves reasoning tasks with aggregation~\citep{wang2023selfconsistency}, verification~\citep{cobbe2021verifiers,genrm,mindthegap}, and search~\citep{yao2023treeofthoughts,hao2023rap}.
Process reward models evaluate intermediate steps~\citep{lightman2023verify}, including through generative verification~\citep{khalifa2026thinkprm,rethinkingrewardmodels}, while $V_1$ uses pairwise comparisons and tournaments to verify mathematical and code solutions~\citep{V1}.
Although most works focus on reasoning tasks, our work mainly addresses agentic tasks and action scaling where each action affects the environment.

\paragraph{Agent test-time scaling.}
Parallel verification~\citep{zhu2025atts,kwok2026llmverifier} and sequential refinement~\citep{scalingtesttimeagenticcoding,SR} scale complete trajectories.
LLM-as-a-Verifier also verifies per-step candidates on Terminal-Bench~\citep{kwok2026llmverifier}.
\citet{guidedsearch} combine action ranking by a trained critic with trajectory selection for SWE tasks.
We instead study verification mechanisms using the generator as verifier across broader terminal tasks.
Other methods evaluate completed patches~\citep{swerm,agenticrubrics} or provide feedback during execution on SWE tasks~\citep{sweprm}.

\section{Conclusion}

We study action scaling for long-horizon terminal agents with the generator and harness fixed.
Useful alternatives from the same generator can improve trajectory success, but wider sampling yields little benefit under weak verification.
Pairwise comparison performs best among the evaluated mechanisms, and distilling the frontier model's responses further improves success without changing the generator.
These gains extend across agent settings, while composition with trajectory scaling reaches higher success at lower estimated token cost than generating more trajectories alone.
These findings establish action scaling as a complementary axis of test-time compute scaling.

\paragraph{Limitations and Future Work.}
Distillation leaves a substantial gap to frontier verification, motivating better training algorithms such as reinforcement learning for verifiers~\citep{V1}.
Our evaluation lacks gold action labels, limiting direct measurement of verification correctness and candidate coverage.
\appautoref{app:limitations} discusses these limitations and future directions in detail.

\subsection*{AI use statement}

In this work, we used generative AI tools to generate synthetic datasets, implement methods, design or provide feedback on research methodology or experiments, support qualitative and thematic data analysis, and assist with translation.
We did not use generative AI tools to propose or refine hypotheses, clean or reformat datasets, or interpret results. The following uses are not applicable to this work: helping develop theoretical models or conceptual frameworks, formulating mathematical claims, providing critical ingredients for proving mathematical claims, and assisting in the writing of proofs.
Additionally, we used generative AI tools to create or modify scientific figures or images and to edit the research paper to improve readability. We reviewed all AI-assisted work. LLM-generated code was checked and tested for correctness by two authors.
We take responsibility for the final content of this work, including text, claims, and artifacts produced with the aid of generative AI.

\subsection*{Ethics statement}

Our experiments evaluate terminal agents in isolated benchmark environments. The action reliability studied here concerns successful task completion and does not establish the safety or security of generated commands. Improvements in terminal-agent capabilities may benefit legitimate automation but could also facilitate harmful activities. Deployment should therefore retain safeguards such as restricted permissions, environment isolation, and human approval for sensitive or irreversible actions. Action verification should complement these safeguards, not replace them.

\subsection*{Reproducibility statement}

We document the verification mechanisms, prompts, and operating settings in \appautoref{app:protocols}, and the distillation data construction and training configuration in \appautoref{app:training}.
\appautoref{app:evaluation} specifies the evaluation tasks, metrics, baseline procedures, and timeout settings.
Inference cost calculations are described in \appautoref{app:compute}, while \appautoref{app:offline} and \appautoref{app:residual} detail the offline agreement analysis and disagreement categorization.

\bibliography{references}

@article{cobbe2021verifiers,
  author       = {Karl Cobbe and
                  Vineet Kosaraju and
                  Mohammad Bavarian and
                  Mark Chen and
                  Heewoo Jun and
                  Lukasz Kaiser and
                  Matthias Plappert and
                  Jerry Tworek and
                  Jacob Hilton and
                  Reiichiro Nakano and
                  Christopher Hesse and
                  John Schulman},
  title        = {Training Verifiers to Solve Math Word Problems},
  journal      = {arXiv},
  volume       = {2110.14168},
  year         = {2021},
  url          = {https://arxiv.org/abs/2110.14168}
}

@inproceedings{wang2023selfconsistency,
  author       = {Xuezhi Wang and
                  Jason Wei and
                  Dale Schuurmans and
                  Quoc V. Le and
                  Ed H. Chi and
                  Sharan Narang and
                  Aakanksha Chowdhery and
                  Denny Zhou},
  title        = {Self-Consistency Improves Chain of Thought Reasoning in Language Models},
  booktitle    = {The Eleventh International Conference on Learning Representations,
                  {ICLR} 2023, Kigali, Rwanda, May 1-5, 2023},
  publisher    = {OpenReview.net},
  year         = {2023},
  url          = {https://openreview.net/forum?id=1PL1NIMMrw}
}

@inproceedings{
snell2024scaling,
title={Scaling {LLM} Test-Time Compute Optimally Can be More Effective than Scaling Parameters for Reasoning},
author={Charlie Victor Snell and Jaehoon Lee and Kelvin Xu and Aviral Kumar},
booktitle={The Thirteenth International Conference on Learning Representations},
year={2025},
url={https://openreview.net/forum?id=4FWAwZtd2n}
}

@article{brown2024monkeys,
  author       = {Bradley C. A. Brown and
                  Jordan Juravsky and
                  Ryan Ehrlich and
                  Ronald Clark and
                  Quoc V. Le and
                  Christopher R{\'{e}} and
                  Azalia Mirhoseini},
  title        = {Large Language Monkeys: Scaling Inference Compute with Repeated Sampling},
  journal      = {arXiv},
  volume       = {2407.21787},
  year         = {2024},
  url          = {https://doi.org/10.48550/arXiv.2407.21787}
}

@inproceedings{lightman2023verify,
  author       = {Hunter Lightman and
                  Vineet Kosaraju and
                  Yuri Burda and
                  Harrison Edwards and
                  Bowen Baker and
                  Teddy Lee and
                  Jan Leike and
                  John Schulman and
                  Ilya Sutskever and
                  Karl Cobbe},
  title        = {Let's Verify Step by Step},
  booktitle    = {The Twelfth International Conference on Learning Representations,
                  {ICLR} 2024, Vienna, Austria, May 7-11, 2024},
  publisher    = {OpenReview.net},
  year         = {2024},
  url          = {https://openreview.net/forum?id=v8L0pN6EOi}
}

@article{khalifa2026thinkprm,
title={Process Reward Models That Think},
author={Khalifa, Muhammad and Agarwal, Rishabh and Logeswaran, Lajanugen and Kim, Jaekyeom and Peng, Hao and Lee, Moontae and Lee, Honglak and Wang, Lu},
journal={Transactions on Machine Learning Research},
year={2026},
url={https://openreview.net/forum?id=FPVCb0WMuN}
}

@inproceedings{yao2023treeofthoughts,
title={Tree of Thoughts: Deliberate Problem Solving with Large Language Models},
author={Yao, Shunyu and Yu, Dian and Zhao, Jeffrey and Shafran, Izhak and Griffiths, Tom and Cao, Yuan and Narasimhan, Karthik},
booktitle={Advances in Neural Information Processing Systems},
volume={36},
year={2023},
url={https://proceedings.neurips.cc/paper/2023/hash/271db9922b8d1f4dd7aaef84ed5ac703-Abstract.html}
}

@inproceedings{hao2023rap,
title={Reasoning with Language Model is Planning with World Model},
author={Hao, Shibo and Gu, Yi and Ma, Haodi and Hong, Joshua and Wang, Zhen and Wang, Daisy and Hu, Zhiting},
booktitle={Proceedings of the 2023 Conference on Empirical Methods in Natural Language Processing},
pages={8154--8173},
year={2023},
publisher={Association for Computational Linguistics},
doi={10.18653/v1/2023.emnlp-main.507},
url={https://aclanthology.org/2023.emnlp-main.507/}
}

@article{kwok2026llmverifier,
  author       = {Jacky Kwok and
                  Shulu Li and
                  Pranav Atreya and
                  Yuejiang Liu and
                  Yixing Jiang and
                  Chelsea Finn and
                  Marco Pavone and
                  Ion Stoica and
                  Azalia Mirhoseini},
  title        = {LLM-as-a-Verifier: {A} General-Purpose Verification Framework},
  journal      = {arXiv},
  volume       = {2607.05391},
  year         = {2026},
  url          = {https://doi.org/10.48550/arXiv.2607.05391}
}

@article{zhu2025atts,
  author       = {King Zhu and
                  Hanhao Li and
                  Siwei Wu and
                  Tianshun Xing and
                  Dehua Ma and
                  Xiangru Tang and
                  Minghao Liu and
                  Jian Yang and
                  Jiaheng Liu and
                  Yuchen Eleanor Jiang and
                  Changwang Zhang and
                  Chenghua Lin and
                  Jun Wang and
                  Ge Zhang and
                  Wangchunshu Zhou},
  title        = {Scaling Test-time Compute for {LLM} Agents},
  journal      = {arXiv},
  volume       = {2506.12928},
  year         = {2025},
  url          = {https://doi.org/10.48550/arXiv.2506.12928}
}

@inproceedings{guidedsearch,
title={Guided Search Strategies in Non-Serializable Environments with Applications to Software Engineering Agents},
author={Zainullina, Karina and Golubev, Alexander and Trofimova, Maria and Polezhaev, Sergei and Badertdinov, Ibragim and Litvintseva, Daria and Karasik, Simon and Fisin, Filipp and Skvortsov, Sergei and Nekrashevich, Maksim and Shevtsov, Anton and Yangel, Boris},
booktitle={International Conference on Machine Learning},
year={2025},
url={https://arxiv.org/abs/2505.13652}
}

@article{egss,
title={{EGSS}: Entropy-guided Stepwise Scaling for Reliable Software Engineering},
author={Mao, Chenhui and Lei, Yuanting and Wei, Zhixiang and Liang, Ming and Wang, Zhixiang and Xu, Jingxuan and Chen, Dajun and Jiang, Wei and Li, Yong},
journal={arXiv},
volume={2602.05242},
year={2026},
url={https://arxiv.org/abs/2602.05242}
}

@misc{tbench2025,
title={Terminal-Bench: A Benchmark for {AI} Agents in Terminal Environments},
author={{The Terminal-Bench Team}},
howpublished={\url{https://github.com/laude-institute/terminal-bench}},
year={2025}
}

@inproceedings{jimenez2024swebench,
title={{SWE}-bench: Can Language Models Resolve Real-World {GitHub} Issues?},
author={Jimenez, Carlos E. and Yang, John and Wettig, Alexander and Yao, Shunyu and Pei, Kexin and Press, Ofir and Narasimhan, Karthik},
booktitle={International Conference on Learning Representations},
year={2024}
}

@inproceedings{yao2023react,
title={{ReAct}: Synergizing Reasoning and Acting in Language Models},
author={Yao, Shunyu and Zhao, Jeffrey and Yu, Dian and Du, Nan and Shafran, Izhak and Narasimhan, Karthik and Cao, Yuan},
booktitle={International Conference on Learning Representations},
year={2023}
}

@article{nemotronterminal,
  author       = {Renjie Pi and
                  Grace Lam and
                  Mohammad Shoeybi and
                  Pooya Jannaty and
                  Bryan Catanzaro and
                  Wei Ping},
  title        = {On Data Engineering for Scaling {LLM} Terminal Capabilities},
  journal      = {arXiv},
  volume       = {2602.21193},
  year         = {2026},
  url          = {https://doi.org/10.48550/arXiv.2602.21193},
}

@article{endlessterminals,
      title={Endless Terminals: Scaling RL Environments for Terminal Agents}, 
      author={Kanishk Gandhi and Shivam Garg and Noah D. Goodman and Dimitris Papailiopoulos},
      year={2026},
      journal={arXiv},
      volume={2601.16443},
      url={https://arxiv.org/abs/2601.16443}, 
}

@article{openthoughts,
      title={OpenThoughts-Agent: Data Recipes for Agentic Models}, 
      author={Negin Raoof and Richard Zhuang and Marianna Nezhurina and Etash Guha and Atula Tejaswi and Ryan Marten and Charlie F. Ruan and Tyler Griggs and Alexander Glenn Shaw and Hritik Bansal and E. Kelly Buchanan and Artem Gazizov and Reinhard Heckel and Chinmay Hegde and Sankalp Jajee and Daanish Khazi and Emmanouil Koukoumidis and Xiangyi Li and Hange Liu and Shlok Natarajan and Harsh Raj and Nicholas Roberts and Ethan Shen and Nishad Singhi and Michael Siu and Ashima Suvarna and Hanwen Xing and Patrick Yubeaton and Robert Zhang and Leon Liangyu Chen and Xiaokun Chen and Steven Dillmann and Saadia Gabriel and Xunyi Jiang and Anurag Kashyap and Boxuan Li and Yein Park and Minh Pham and Sujay Sanghavi and Lin Shi and Ke Sun and Yixin Wang and Zhiwei Xu and Erica Zhang and Siyan Zhao and Wanjia Zhao and Jenia Jitsev and Alex Dimakis and Benjamin Feuer and Ludwig Schmidt},
      year={2026},
      journal={arXiv},
      volume={2606.24855},
      url={https://arxiv.org/abs/2606.24855}, 
}

@article{tmax,
      title={Tmax: A simple recipe for terminal agents}, 
      author={Hamish Ivison and Junjie Oscar Yin and Rulin Shao and Teng Xiao and Nathan Lambert and Hannaneh Hajishirzi},
      year={2026},
      journal={arXiv},
      volume={2606.23321},
      url={https://arxiv.org/abs/2606.23321}, 
}

@inproceedings{s1,
  author       = {Niklas Muennighoff and
                  Zitong Yang and
                  Weijia Shi and
                  Xiang Lisa Li and
                  Li Fei{-}Fei and
                  Hannaneh Hajishirzi and
                  Luke Zettlemoyer and
                  Percy Liang and
                  Emmanuel J. Cand{\`{e}}s and
                  Tatsunori Hashimoto},
  editor       = {Christos Christodoulopoulos and
                  Tanmoy Chakraborty and
                  Carolyn Rose and
                  Violet Peng},
  title        = {s1: Simple test-time scaling},
  booktitle    = {Proceedings of the 2025 Conference on Empirical Methods in Natural
                  Language Processing, {EMNLP} 2025, Suzhou, China, November 4-9, 2025},
  pages        = {20275--20321},
  publisher    = {Association for Computational Linguistics},
  year         = {2025},
  url          = {https://doi.org/10.18653/v1/2025.emnlp-main.1025}
}

@article{scalingtesttimeagenticcoding,
  author       = {Joongwon Kim and
                  Wannan Yang and
                  Kelvin Niu and
                  Hongming Zhang and
                  Yun Zhu and
                  Eryk Helenowski and
                  Ruan Silva and
                  Zhengxing Chen and
                  Srinivasan Iyer and
                  Manzil Zaheer and
                  Daniel Fried and
                  Hannaneh Hajishirzi and
                  Sanjeev Arora and
                  Gabriel Synnaeve and
                  Ruslan Salakhutdinov and
                  Anirudh Goyal},
  title        = {Scaling Test-Time Compute for Agentic Coding},
  journal      = {arXiv},
  volume       = {2604.16529},
  year         = {2026},
  url          = {https://doi.org/10.48550/arXiv.2604.16529}
}

@inproceedings{
genrm,
title={Generative Verifiers: Reward Modeling as Next-Token Prediction},
author={Lunjun Zhang and Arian Hosseini and Hritik Bansal and Mehran Kazemi and Aviral Kumar and Rishabh Agarwal},
booktitle={The Thirteenth International Conference on Learning Representations},
year={2025},
url={https://openreview.net/forum?id=Ccwp4tFEtE}
}

@article{rethinkingrewardmodels,
  author       = {Dong Bok Lee and
                  Seanie Lee and
                  Sangwoo Park and
                  Minki Kang and
                  Jinheon Baek and
                  Dongki Kim and
                  Dominik Wagner and
                  Jiongdao Jin and
                  Heejun Lee and
                  Tobias Bocklet and
                  Jinyu Wang and
                  Jingjing Fu and
                  Sung Ju Hwang and
                  Jiang Bian and
                  Lei Song},
  title        = {Rethinking Reward Models for Multi-Domain Test-Time Scaling},
  journal      = {Trans. Mach. Learn. Res.},
  volume       = {2026},
  year         = {2026},
  url          = {https://openreview.net/forum?id=PgouBhL7IR}
}

@inproceedings{
mindthegap,
title={Mind the Gap: Examining the Self-Improvement Capabilities of Large Language Models},
author={Yuda Song and Hanlin Zhang and Carson Eisenach and Sham M. Kakade and Dean Foster and Udaya Ghai},
booktitle={The Thirteenth International Conference on Learning Representations},
year={2025},
url={https://openreview.net/forum?id=mtJSMcF3ek}
}

@article{swerm,
  author       = {KaShun Shum and
                  Binyuan Hui and
                  Jiawei Chen and
                  Lei Zhang and
                  X. W. and
                  Jiaxi Yang and
                  Yuzhen Huang and
                  Junyang Lin and
                  Junxian He},
  title        = {{SWE-RM:} Execution-free Feedback For Software Engineering Agents},
  journal      = {arXiv},
  volume       = {2512.21919},
  year         = {2025},
  url          = {https://doi.org/10.48550/arXiv.2512.21919},
}

@inproceedings{agenticrubrics,
  author       = {Mohit Raghavendra and
                  Anisha Gunjal and
                  Bing Liu and
                  Yunzhong He},
  editor       = {Maria Liakata and
                  Viviane P. Moreira and
                  Jiajun Zhang and
                  David Jurgens},
  title        = {Agentic Rubrics as Contextual Verifiers for {SWE} Agents},
  booktitle    = {Proceedings of the 64th Annual Meeting of the Association for Computational
                  Linguistics (Volume 1: Long Papers), {ACL} 2026, San Diego, California,
                  United States, July 2-7, 2026},
  pages        = {15265--15290},
  publisher    = {Association for Computational Linguistics},
  year         = {2026},
  url          = {https://doi.org/10.18653/v1/2026.acl-long.697}
}

@article{V1,
      title={$V_1$: Unifying Generation and Self-Verification for Parallel Reasoners}, 
      author={Harman Singh and Xiuyu Li and Kusha Sareen and Monishwaran Maheswaran and Sijun Tan and Xiaoxia Wu and Junxiong Wang and Alpay Ariyak and Qingyang Wu and Samir Khaki and Rishabh Tiwari and Long Lian and Yucheng Lu and Boyi Li and Alane Suhr and Ben Athiwaratkun and Kurt Keutzer},
      year={2026},
      journal={arXiv},
      volume={2603.04304},
      url={https://arxiv.org/abs/2603.04304}, 
}

@article{RecoveryBench,
  author = {Letta},
  title = {Introducing Recovery-Bench: Evaluating LLMs' Ability to Recover from Mistakes},
  journal = {Letta Blog},
  year = {2025},
  month = {August},
  url = {https://www.letta.com/blog/recovery-bench/},
}

@article{DeepseekR1,
   title={DeepSeek-R1 incentivizes reasoning in LLMs through reinforcement learning},
   volume={645},
   ISSN={1476-4687},
   url={http://dx.doi.org/10.1038/s41586-025-09422-z},
   DOI={10.1038/s41586-025-09422-z},
   number={8081},
   journal={Nature},
   publisher={Springer Science and Business Media LLC},
   author={Guo, Daya and Yang, Dejian and Zhang, Haowei and Song, Junxiao and Wang, Peiyi and Zhu, Qihao and Xu, Runxin and Zhang, Ruoyu and Ma, Shirong and Bi, Xiao and Zhang, Xiaokang and Yu, Xingkai and Wu, Yu and Wu, Z. F. and Gou, Zhibin and Shao, Zhihong and Li, Zhuoshu and Gao, Ziyi and Liu, Aixin and Xue, Bing and Wang, Bingxuan and Wu, Bochao and Feng, Bei and Lu, Chengda and Zhao, Chenggang and Deng, Chengqi and Ruan, Chong and Dai, Damai and Chen, Deli and Ji, Dongjie and Li, Erhang and Lin, Fangyun and Dai, Fucong and Luo, Fuli and Hao, Guangbo and Chen, Guanting and Li, Guowei and Zhang, H. and Xu, Hanwei and Ding, Honghui and Gao, Huazuo and Qu, Hui and Li, Hui and Guo, Jianzhong and Li, Jiashi and Chen, Jingchang and Yuan, Jingyang and Tu, Jinhao and Qiu, Junjie and Li, Junlong and Cai, J. L. and Ni, Jiaqi and Liang, Jian and Chen, Jin and Dong, Kai and Hu, Kai and You, Kaichao and Gao, Kaige and Guan, Kang and Huang, Kexin and Yu, Kuai and Wang, Lean and Zhang, Lecong and Zhao, Liang and Wang, Litong and Zhang, Liyue and Xu, Lei and Xia, Leyi and Zhang, Mingchuan and Zhang, Minghua and Tang, Minghui and Zhou, Mingxu and Li, Meng and Wang, Miaojun and Li, Mingming and Tian, Ning and Huang, Panpan and Zhang, Peng and Wang, Qiancheng and Chen, Qinyu and Du, Qiushi and Ge, Ruiqi and Zhang, Ruisong and Pan, Ruizhe and Wang, Runji and Chen, R. J. and Jin, R. L. and Chen, Ruyi and Lu, Shanghao and Zhou, Shangyan and Chen, Shanhuang and Ye, Shengfeng and Wang, Shiyu and Yu, Shuiping and Zhou, Shunfeng and Pan, Shuting and Li, S. S. and Zhou, Shuang and Wu, Shaoqing and Yun, Tao and Pei, Tian and Sun, Tianyu and Wang, T. and Zeng, Wangding and Liu, Wen and Liang, Wenfeng and Gao, Wenjun and Yu, Wenqin and Zhang, Wentao and Xiao, W. L. and An, Wei and Liu, Xiaodong and Wang, Xiaohan and Chen, Xiaokang and Nie, Xiaotao and Cheng, Xin and Liu, Xin and Xie, Xin and Liu, Xingchao and Yang, Xinyu and Li, Xinyuan and Su, Xuecheng and Lin, Xuheng and Li, X. Q. and Jin, Xiangyue and Shen, Xiaojin and Chen, Xiaosha and Sun, Xiaowen and Wang, Xiaoxiang and Song, Xinnan and Zhou, Xinyi and Wang, Xianzu and Shan, Xinxia and Li, Y. K. and Wang, Y. Q. and Wei, Y. X. and Zhang, Yang and Xu, Yanhong and Li, Yao and Zhao, Yao and Sun, Yaofeng and Wang, Yaohui and Yu, Yi and Zhang, Yichao and Shi, Yifan and Xiong, Yiliang and He, Ying and Piao, Yishi and Wang, Yisong and Tan, Yixuan and Ma, Yiyang and Liu, Yiyuan and Guo, Yongqiang and Ou, Yuan and Wang, Yuduan and Gong, Yue and Zou, Yuheng and He, Yujia and Xiong, Yunfan and Luo, Yuxiang and You, Yuxiang and Liu, Yuxuan and Zhou, Yuyang and Zhu, Y. X. and Huang, Yanping and Li, Yaohui and Zheng, Yi and Zhu, Yuchen and Ma, Yunxian and Tang, Ying and Zha, Yukun and Yan, Yuting and Ren, Z. Z. and Ren, Zehui and Sha, Zhangli and Fu, Zhe and Xu, Zhean and Xie, Zhenda and Zhang, Zhengyan and Hao, Zhewen and Ma, Zhicheng and Yan, Zhigang and Wu, Zhiyu and Gu, Zihui and Zhu, Zijia and Liu, Zijun and Li, Zilin and Xie, Ziwei and Song, Ziyang and Pan, Zizheng and Huang, Zhen and Xu, Zhipeng and Zhang, Zhongyu and Zhang, Zhen},
   year={2025},
   month=sep, pages={633–638} }

@inproceedings{
    webshepherd,
    title={Web-Shepherd: Advancing {PRM}s for Reinforcing Web Agents},
    author={Hyungjoo Chae and Sunghwan Kim and Junhee Cho and Seungone Kim and Seungjun Moon and Gyeom Hwangbo and Dongha Lim and Minjin Kim and Yeonjun Hwang and Minju Gwak and Dongwook Choi and Minseok Kang and Gwanhoon Im and ByeongUng Cho and Hyojun Kim and Jun Hee Han and Taeyoon Kwon and Minju Kim and Beong-woo Kwak and Dongjin Kang and Jinyoung Yeo},
    booktitle={The Thirty-ninth Annual Conference on Neural Information Processing Systems},
    year={2025},
    url={https://openreview.net/forum?id=G2kMroO9UV}
}

@misc{TBLite,
  author = {OpenThoughts-Agent team and Snorkel AI and Bespoke Labs},
  month = Feb,
  title = {{OpenThoughts-TBLite: A High-Signal Benchmark for Iterating on Terminal Agents}},
  howpublished = {https://www.openthoughts.ai/blog/openthoughts-tblite},
  year = {2026}
}

@misc{GPT5.6,
  author = {OpenAI},
  title = {GPT-5.6: Frontier intelligence that scales with your ambition},
  howpublished = {\url{https://openai.com/index/gpt-5-6/}},
  year={2026},
}

@article{GenSelect,
  author       = {Shubham Toshniwal and
                  Ivan Sorokin and
                  Aleksander Ficek and
                  Ivan Moshkov and
                  Igor Gitman},
  title        = {GenSelect: {A} Generative Approach to Best-of-N},
  journal      = {arXiv},
  volume       = {2507.17797},
  year         = {2025},
  url          = {https://doi.org/10.48550/arXiv.2507.17797}
}

@inproceedings{swe-agent,
  title={{SWE}-agent: Agent-Computer Interfaces Enable Automated Software Engineering},
  author={John Yang and Carlos E Jimenez and Alexander Wettig and Kilian Lieret and Shunyu Yao and Karthik R Narasimhan and Ofir Press},
  booktitle={The Thirty-eighth Annual Conference on Neural Information Processing Systems},
  year={2024},
  url={https://arxiv.org/abs/2405.15793}
}

@inproceedings{
acon,
title={{ACON}: Optimizing Context Compression for Long-horizon {LLM} Agents},
author={Minki Kang and Wei-Ning Chen and Dongge Han and Huseyin A Inan and Lukas Wutschitz and Yanzhi Chen and Robert Sim and Saravan Rajmohan},
booktitle={Forty-third International Conference on Machine Learning},
year={2026},
url={https://openreview.net/forum?id=5EmOOLtH5P}
}

@article{preping,
      title={PREPING: Building Agent Memory without Tasks}, 
      author={Yumin Choi and Sangwoo Park and Minki Kang and Jinheon Baek and Sung Ju Hwang},
      year={2026},
      journal={arXiv},
      volume={2605.13880},
      url={https://arxiv.org/abs/2605.13880}, 
}

@article{primeagent,
      title={Prime Agent: A Self-Improving RLM Harness}, 
      author={Seth Karten and Alex L. Zhang and Kevin Thomas and Sebastian Müller and Elie Bakouch and Daniel Auras and Mika Senghaas and Fares Obeid and Konstantin Dunas and Johannes Hagemann and Sami Jaghouar},
      year={2026},
      journal={arXiv},
      volume={2608.23552},
      url={https://arxiv.org/abs/2608.23552}, 
}

@inproceedings{
openhands,
title={OpenHands: An Open Platform for {AI} Software Developers as Generalist Agents},
author={Xingyao Wang and Boxuan Li and Yufan Song and Frank F. Xu and Xiangru Tang and Mingchen Zhuge and Jiayi Pan and Yueqi Song and Bowen Li and Jaskirat Singh and Hoang H. Tran and Fuqiang Li and Ren Ma and Mingzhang Zheng and Bill Qian and Daniel Shao and Niklas Muennighoff and Yizhe Zhang and Binyuan Hui and Junyang Lin and Robert Brennan and Hao Peng and Heng Ji and Graham Neubig},
booktitle={The Thirteenth International Conference on Learning Representations},
year={2025},
url={https://openreview.net/forum?id=OJd3ayDDoF}
}

@article{meta-harness,
      title={Meta-Harness: End-to-End Optimization of Model Harnesses}, 
      author={Yoonho Lee and Roshen Nair and Qizheng Zhang and Kangwook Lee and Omar Khattab and Chelsea Finn},
      year={2026},
      journal={arXiv},
      volume={2603.28052},
      url={https://arxiv.org/abs/2603.28052}, 
}

@inproceedings{hu2021lora,
  author       = {Edward J. Hu and
                  Yelong Shen and
                  Phillip Wallis and
                  Zeyuan Allen{-}Zhu and
                  Yuanzhi Li and
                  Shean Wang and
                  Lu Wang and
                  Weizhu Chen},
  title        = {LoRA: Low-Rank Adaptation of Large Language Models},
  booktitle    = {The Tenth International Conference on Learning Representations, {ICLR}
                  2022, Virtual Event, April 25-29, 2022},
  publisher    = {OpenReview.net},
  year         = {2022},
  url          = {https://openreview.net/forum?id=nZeVKeeFYf9}
}

@article{sweprm,
  author       = {Shubham Gandhi and
                  Jason Tsay and
                  Jatin Ganhotra and
                  Kiran Kate and
                  Yara Rizk},
  title        = {When Agents go Astray: Course-Correcting {SWE} Agents with PRMs},
  journal      = {arXiv},
  volume       = {2509.02360},
  year         = {2025},
  url          = {https://doi.org/10.48550/arXiv.2509.02360}
}

@article{criu,
  author       = {Fabio Andrijauskas and
                  Igor Sfiligoi and
                  Diego Davila and
                  Aashay Arora and
                  Jonathan Guiang and
                  Brian Bockelman and
                  Greg Thain and
                  Frank W{\"{u}}rthwein},
  title        = {{CRIU} - Checkpoint Restore in Userspace for computational simulations
                  and scientific applications},
  journal      = {arXiv},
  volume       = {2402.05244},
  year         = {2024},
  url          = {https://doi.org/10.48550/arXiv.2402.05244},
}

@article{SR,
  author       = {Lovish Madaan and
                  Aniket Didolkar and
                  Suchin Gururangan and
                  John Quan and
                  Ruan Silva and
                  Ruslan Salakhutdinov and
                  Manzil Zaheer and
                  Sanjeev Arora and
                  Anirudh Goyal},
  title        = {Rethinking Thinking Tokens: LLMs as Improvement Operators},
  journal      = {arXiv},
  volume       = {2510.01123},
  year         = {2025},
  url          = {https://doi.org/10.48550/arXiv.2510.01123}
}

@misc{qwen3.5,
  author = {QwenTeam},
  title = {Qwen3.5: Towards Native Multimodal Agents},
  howpublished = {\url{https://qwen.ai/blog?id=qwen3.5}},
  year={2026},
}

@inproceedings{
featurebench,
title={FeatureBench: Benchmarking Agentic Coding for Complex Feature Development},
author={Qixing Zhou and JiaCheng Zhang and Haiyang Wang and Rui Hao and Jiahe Wang and Minghao Han and Yuxue Yang and Shuzhe Wu and Feiyang Pan and Lue Fan and Dandan Tu and Zhaoxiang Zhang},
booktitle={The Fourteenth International Conference on Learning Representations},
year={2026},
url={https://openreview.net/forum?id=41xrZ3uGuI}
}

@inproceedings{skd,
  author       = {Yoon Kim and
                  Alexander M. Rush},
  editor       = {Jian Su and
                  Kevin Duh and
                  Xavier Carreras},
  title        = {Sequence-Level Knowledge Distillation},
  booktitle    = {Proceedings of the 2016 Conference on Empirical Methods in Natural
                  Language Processing, {EMNLP} 2016, Austin, Texas, USA, November 1-4,
                  2016},
  pages        = {1317--1327},
  publisher    = {The Association for Computational Linguistics},
  year         = {2016},
  url          = {https://doi.org/10.18653/v1/d16-1139},
}

@article{nemo-ultra,
  author       = {NVIDIA},
  title        = {Nemotron 3 Ultra: Open, Efficient Mixture-of-Experts Hybrid Mamba-Transformer
                  Model for Agentic Reasoning},
  journal      = {arXiv},
  volume       = {2606.15007},
  year         = {2026},
  url          = {https://doi.org/10.48550/arXiv.2606.15007},
}

@misc{nemo-lightning,
  author = {NVIDIA},
  title = {NVIDIA Nemotron 3.5 Lightning Delivers Fast, Accurate Specialized Task Execution for Long-Running Agents},
  howpublished = {\url{https://developer.nvidia.com/blog/nvidia-nemotron-3-5-lightning-delivers-fast-accurate-specialized-task-execution-for-long-running-agents/}},
  year={2026},
}

@inproceedings{mathsheperd,
  author       = {Peiyi Wang and
                  Lei Li and
                  Zhihong Shao and
                  Runxin Xu and
                  Damai Dai and
                  Yifei Li and
                  Deli Chen and
                  Yu Wu and
                  Zhifang Sui},
  editor       = {Lun{-}Wei Ku and
                  Andre Martins and
                  Vivek Srikumar},
  title        = {Math-Shepherd: Verify and Reinforce LLMs Step-by-step without Human
                  Annotations},
  booktitle    = {Proceedings of the 62nd Annual Meeting of the Association for Computational
                  Linguistics (Volume 1: Long Papers), {ACL} 2024, Bangkok, Thailand,
                  August 11-16, 2024},
  pages        = {9426--9439},
  publisher    = {Association for Computational Linguistics},
  year         = {2024},
  url          = {https://doi.org/10.18653/v1/2024.acl-long.510}
}

@inproceedings{processbench,
  author       = {Chujie Zheng and
                  Zhenru Zhang and
                  Beichen Zhang and
                  Runji Lin and
                  Keming Lu and
                  Bowen Yu and
                  Dayiheng Liu and
                  Jingren Zhou and
                  Junyang Lin},
  editor       = {Wanxiang Che and
                  Joyce Nabende and
                  Ekaterina Shutova and
                  Mohammad Taher Pilehvar},
  title        = {ProcessBench: Identifying Process Errors in Mathematical Reasoning},
  booktitle    = {Proceedings of the 63rd Annual Meeting of the Association for Computational
                  Linguistics (Volume 1: Long Papers), {ACL} 2025, Vienna, Austria,
                  July 27 - August 1, 2025},
  pages        = {1009--1024},
  publisher    = {Association for Computational Linguistics},
  year         = {2025},
  url          = {https://doi.org/10.18653/v1/2025.acl-long.50},
}

@inproceedings{inferencescalinglaws,
  author       = {Yangzhen Wu and
                  Zhiqing Sun and
                  Shanda Li and
                  Sean Welleck and
                  Yiming Yang},
  title        = {Inference Scaling Laws: An Empirical Analysis of Compute-Optimal Inference
                  for {LLM} Problem-Solving},
  booktitle    = {The Thirteenth International Conference on Learning Representations,
                  {ICLR} 2025, Singapore, April 24-28, 2025},
  publisher    = {OpenReview.net},
  year         = {2025},
  url          = {https://openreview.net/forum?id=VNckp7JEHn}
}

@article{qwen-agentworld,
  author       = {Yuxin Zuo and
                  Zikai Xiao and
                  Li Sheng and
                  Fei Huang and
                  Jianhong Tu and
                  Yuxuan Liu and
                  Tianyi Tang and
                  Xiaomeng Hu and
                  Yang Su and
                  Qingfeng Lan and
                  Yantao Liu and
                  Qin Zhu and
                  Yinger Zhang and
                  Bowen Yu and
                  Haiquan Zhao and
                  Haiyang Xu and
                  Jianxin Yang and
                  Jiayang Cheng and
                  Junyang Wang and
                  Lianghao Deng and
                  Mingfeng Xue and
                  Tianyi Bai and
                  Yang Fan and
                  Yubo Ma and
                  Yucheng Li and
                  Zeyu Cui and
                  Zhihai Wang and
                  Zhihui Xie and
                  Zhuorui Ye and
                  An Yang and
                  Dayiheng Liu and
                  Jingren Zhou and
                  Ning Ding},
  title        = {Qwen-AgentWorld: Language World Models for General Agents},
  journal      = {arXiv},
  volume       = {2606.24597},
  year         = {2026},
  url          = {https://doi.org/10.48550/arXiv.2606.24597}
}

@article{merrill2026terminalbench,
  title={Terminal-Bench: Benchmarking Agents on Hard, Realistic Tasks in Command Line Interfaces},
  author={Merrill, Mike A. and Shaw, Alexander G. and Carlini, Nicholas and Li, Boxuan and Raj, Harsh and others},
  journal={arXiv},
volume={2601.11868},
  year={2026},
  url={https://arxiv.org/abs/2601.11868}
}

@misc{jev,
  author = {Diogo Almeida},
  title = {Introducing System One Models \& {Jev}},
  year = {2026},
  month = sep,
  howpublished = {TypeSafe AI Blog},
  url = {https://typesafe.ai/blog/introducing-system-one-models-and-jev}
}

@article{frognano,
      title={FrogNano: Training a 4B Coding Agent via Online Task Synthesis}, 
      author={Minseon Kim and Zhengyan Shi and Emiliano Penaloza and Christopher Cui and Roger Creus Castanyer and Maryam Hashemzadeh and Isadora White and Jonathan Light and Jeonghye Kim and Matheus Pereira and Darya Moldavskaya and Chinmay Singh and Fabio Vera and Baolin Peng and Xingdi Yuan and Marc-Alexandre Côté and Alessandro Sordoni},
      year={2026},
      journal={arXiv},
      volume={2609.07925},
      url={https://arxiv.org/abs/2609.07925}, 
}

@article{lee2020training,
  title={Training encoder-attention through fully-connected crfs for efficient end-to-end lane detection model},
  author={Lee, Byung-Kwan},
  year={2020},
  publisher={KAIST}
}

@misc{lee2021towards,
title={Towards Adversarial Robustness of Bayesian Neural Network through Hierarchical Variational Inference},
author={Byung-Kwan Lee and Youngjoon Yu and Yong Man Ro},
year={2021},
url={https://openreview.net/forum?id=Cue2ZEBf12}
}

@inproceedings{NEURIPS2021_8e5e15c4,
 author = {Kim, Junho and Lee, Byung-Kwan and Ro, Yong Man},
 booktitle = {Advances in Neural Information Processing Systems},
 editor = {M. Ranzato and A. Beygelzimer and Y. Dauphin and P.S. Liang and J. Wortman Vaughan},
 pages = {17148--17159},
 publisher = {Curran Associates, Inc.},
 title = {Distilling Robust and Non-Robust Features in Adversarial Examples by Information Bottleneck},
 url = {https://proceedings.neurips.cc/paper_files/paper/2021/file/8e5e15c4e6d09c8333a17843461041a9-Paper.pdf},
 volume = {34},
 year = {2021}
}

@InProceedings{Lee_2022_CVPR,
    author    = {Lee, Byung-Kwan and Kim, Junho and Ro, Yong Man},
    title     = {Masking Adversarial Damage: Finding Adversarial Saliency for Robust and Sparse Network},
    booktitle = {Proceedings of the IEEE/CVF Conference on Computer Vision and Pattern Recognition (CVPR)},
    month     = {June},
    year      = {2022},
    pages     = {15126-15136}
}

@InProceedings{Kim_2023_CVPR,
    author    = {Kim, Junho and Lee, Byung-Kwan and Ro, Yong Man},
    title     = {Demystifying Causal Features on Adversarial Examples and Causal Inoculation for Robust Network by Adversarial Instrumental Variable Regression},
    booktitle = {Proceedings of the IEEE/CVF Conference on Computer Vision and Pattern Recognition (CVPR)},
    month     = {June},
    year      = {2023},
    pages     = {12302-12312}
}

@INPROCEEDINGS{10222502,
  author={Kim, Yeonju and Kim, Junho and Lee, Byung-Kwan and Shin, Sebin and Ro, Yong Man},
  booktitle={2023 IEEE International Conference on Image Processing (ICIP)}, 
  title={Mitigating Dataset Bias in Image Captioning Through Clip Confounder-Free Captioning Network}, 
  year={2023},
  volume={},
  number={},
  pages={1720-1724},
  doi={10.1109/ICIP49359.2023.10222502}}

@InProceedings{Lee_2023_ICCV,
    author    = {Lee, Byung-Kwan and Kim, Junho and Ro, Yong Man},
    title     = {Mitigating Adversarial Vulnerability through Causal Parameter Estimation by Adversarial Double Machine Learning},
    booktitle = {Proceedings of the IEEE/CVF International Conference on Computer Vision (ICCV)},
    month     = {October},
    year      = {2023},
    pages     = {4499-4509}
}

@article{KIM2026112173,
title = {Causal unsupervised semantic segmentation},
journal = {Pattern Recognition},
volume = {171},
pages = {112173},
year = {2026},
issn = {0031-3203},
doi = {https://doi.org/10.1016/j.patcog.2025.112173},
url = {https://www.sciencedirect.com/science/article/pii/S0031320325008349},
author = {Junho Kim and Byung-Kwan Lee and Yong Man Ro}
}

@inproceedings{lee-etal-2024-collavo,
    title = "{C}o{LL}a{VO}: Crayon Large Language and Vision m{O}del",
    author = "Lee, Byung-Kwan  and
      Park, Beomchan  and
      Kim, Chae Won  and
      Ro, Yong Man",
    editor = "Ku, Lun-Wei  and
      Martins, Andre  and
      Srikumar, Vivek",
    booktitle = "Findings of the Association for Computational Linguistics: ACL 2024",
    month = aug,
    year = "2024",
    address = "Bangkok, Thailand",
    publisher = "Association for Computational Linguistics",
    url = "https://aclanthology.org/2024.findings-acl.66/",
    doi = "10.18653/v1/2024.findings-acl.66",
    pages = "1121--1138"
}

@InProceedings{10.1007/978-3-031-72967-6_16,
author="Lee, Byung-Kwan
and Park, Beomchan
and Won Kim, Chae
and Man Ro, Yong",
editor="Leonardis, Ale{\v{s}}
and Ricci, Elisa
and Roth, Stefan
and Russakovsky, Olga
and Sattler, Torsten
and Varol, G{\"u}l",
title="MoAI: Mixture of All Intelligence for Large Language and Vision Models",
booktitle="Computer Vision -- ECCV 2024",
year="2025",
publisher="Springer Nature Switzerland",
address="Cham",
pages="273--302",
isbn="978-3-031-72967-6"
}

@inproceedings{NEURIPS2024_473a9a75,
 author = {Lee, Byung-Kwan and Kim, Chae Won and Park, Beomchan and Ro, Yong Man},
 booktitle = {Advances in Neural Information Processing Systems},
 doi = {10.52202/079017-1274},
 editor = {A. Globerson and L. Mackey and D. Belgrave and A. Fan and U. Paquet and J. Tomczak and C. Zhang},
 pages = {40278--40315},
 publisher = {Curran Associates, Inc.},
 title = {Meteor: Mamba-based Traversal of Rationale for Large Language and Vision Models},
 url = {https://proceedings.neurips.cc/paper_files/paper/2024/file/473a9a75edc46eff5ff224d53d5f7294-Paper-Conference.pdf},
 volume = {37},
 year = {2024}
}

@inproceedings{lee-etal-2024-trol,
    title = "{T}ro{L}: Traversal of Layers for Large Language and Vision Models",
    author = "Lee, Byung-Kwan  and
      Chung, Sangyun  and
      Kim, Chae Won  and
      Park, Beomchan  and
      Ro, Yong Man",
    editor = "Al-Onaizan, Yaser  and
      Bansal, Mohit  and
      Chen, Yun-Nung",
    booktitle = "Proceedings of the 2024 Conference on Empirical Methods in Natural Language Processing",
    month = nov,
    year = "2024",
    address = "Miami, Florida, USA",
    publisher = "Association for Computational Linguistics",
    url = "https://aclanthology.org/2024.emnlp-main.633/",
    doi = "10.18653/v1/2024.emnlp-main.633",
    pages = "11314--11342"
}

@article{lee2024phantom,
  title={Phantom of latent for large language and vision models},
  author={Lee, Byung-Kwan and Chung, Sangyun and Kim, Chae Won and Park, Beomchan and Ro, Yong Man},
  journal={arXiv preprint arXiv:2409.14713},
  year={2024}
}

@InProceedings{Lee_2025_CVPR,
    author    = {Lee, Byung-Kwan and Hachiuma, Ryo and Wang, Yu-Chiang Frank and Ro, Yong Man and Wu, Yueh-Hua},
    title     = {VLsI: Verbalized Layers-to-Interactions from Large to Small Vision Language Models},
    booktitle = {Proceedings of the IEEE/CVF Conference on Computer Vision and Pattern Recognition (CVPR)},
    month     = {June},
    year      = {2025},
    pages     = {29545-29557}
}

@misc{wu2026layer,
  title={Layer-wise knowledge distillation for vision language models},
  author={Wu, Yueh-Hua and Lee, Byung-Kwan and Hachiuma, Ryo and Wang, Yu-Chiang},
  year={2026},
  month=may # "~21",
  publisher={Google Patents},
  note={US Patent App. 19/256,885}
}

@InProceedings{Lee_2025_ICCV,
    author    = {Lee, Young-Jun and Lee, Byung-Kwan and Zhang, Jianshu and Hwang, Yechan and Ko, Byungsoo and Kim, Han-Gyu and Yao, Dongyu and Rong, Xuankun and Joo, Eojin and Han, Seung-Ho and Ko, Bowon and Choi, Ho-Jin},
    title     = {MultiVerse: A Multi-Turn Conversation Benchmark for Evaluating Large Vision and Language Models},
    booktitle = {Proceedings of the IEEE/CVF International Conference on Computer Vision (ICCV)},
    month     = {October},
    year      = {2025},
    pages     = {708-719}
}

@inproceedings{NEURIPS2025_e5849736,
 author = {Lee, Byung-Kwan and Hachiuma, Ryo and Ro, Yong Man and Wang, Frank and Wu, Yueh-Hua},
 booktitle = {Advances in Neural Information Processing Systems},
 editor = {D. Belgrave and C. Zhang and H. Lin and R. Pascanu and P. Koniusz and M. Ghassemi and N. Chen},
 pages = {156508--156534},
 publisher = {Curran Associates, Inc.},
 title = {Unified Reinforcement and Imitation Learning for Vision-Language Models},
 url = {https://proceedings.neurips.cc/paper_files/paper/2025/file/e58497367bc8730f61a87d37800c0a06-Paper-Conference.pdf},
 volume = {38},
 year = {2025}
}

@phdthesis{lee2025building,
  title={Building High-performing, Efficient-size Vision Language Models: Merge, Modify, and Distill},
  author={Lee, Byung-Kwan},
  school={Korea Advanced Institute of Science and Technology},
  year={2025}
}

@article{leeenhancing,
  title={Enhancing Conversational Agents with Skill-of-Mind-Infused Large Language Model},
  author={Lee, Young-Jun and Lee, Byung-Kwan and Lee, Dokyong and Oh, Kyeong-Jin and Hwang, Yechan and Choi, Ho-Jin and others}
}

@inproceedings{lee2026refinebench,
title={RefineBench: Evaluating Refinement Capability of Language Models via Checklists},
author={Young-Jun Lee and Seungone Kim and Byung-Kwan Lee and Minkyeong Moon and Yechan Hwang and Jong Myoung Kim and Graham Neubig and Sean Welleck and Ho-Jin Choi},
booktitle={The Fourteenth International Conference on Learning Representations},
year={2026},
url={https://openreview.net/forum?id=GYJFJz9Dy5}
}

@InProceedings{Lee_2026_CVPR_Recursive,
    author    = {Lee, Byung-Kwan and Chee, Youngchae and Ro, Yong Man},
    title     = {Recursive Think-Answer Process for LLMs and VLMs},
    booktitle = {Proceedings of the IEEE/CVF Conference on Computer Vision and Pattern Recognition (CVPR) Findings},
    month     = {June},
    year      = {2026},
    pages     = {9608-9621}
}

@InProceedings{Lee_2026_CVPR_Masters,
    author    = {Lee, Byung-Kwan and Wang, Yu-Chiang Frank and Hachiuma, Ryo},
    title     = {Masking Teacher and Reinforcing Student for Distilling Vision-Language Models},
    booktitle = {Proceedings of the IEEE/CVF Conference on Computer Vision and Pattern Recognition (CVPR)},
    month     = {June},
    year      = {2026},
    pages     = {10126-10141}
}

@InProceedings{10.1007/978-3-032-37281-9_9,
author="Lee, Byung-Kwan
and Hachiuma, Ryo
and Ro, Yong Man
and Wang, Yu-Chiang Frank
and Wu, Yueh-Hua",
editor="Favaro, Paolo
and Kukelova, Zuzana
and Maki, Atsuto
and Rohrbach, Anna
and Schindler, Konrad
and Tombari, Federico",
title="GenRecal: Generation After Recalibration from Large to Small Vision-Language Models",
booktitle="Computer Vision -- ECCV 2026",
year="2026",
publisher="Springer Nature Switzerland",
address="Cham",
pages="152--172",
isbn="978-3-032-37281-9"
}

@InProceedings{10.1007/978-3-032-37281-9_15,
author="Kim, Jiwan
and Kim, Kibum
and Kim, Wonjoong
and Lee, Byung-Kwan
and Park, Chanyoung",
editor="Favaro, Paolo
and Kukelova, Zuzana
and Maki, Atsuto
and Rohrbach, Anna
and Schindler, Konrad
and Tombari, Federico",
title="Why and When Visual Token Pruning Fails? A Study on Relevant Visual Information Shift in MLLMs Decoding",
booktitle="Computer Vision -- ECCV 2026",
year="2026",
publisher="Springer Nature Switzerland",
address="Cham",
pages="244--263",
isbn="978-3-032-37281-9"
}

@article{yu2026hide,
  title={Hide to See: Reasoning-prefix Masking for Visual-anchored Thinking in VLM Distillation},
  author={Yu, Seonghoon and Nam, Dongjun and Lee, Byung-Kwan and Son, Jeany},
  journal={arXiv preprint arXiv:2605.11651},
  year={2026}
}

@article{kang2026agent,
  title={Agent Explorative Policy Optimization for Multimodal Agentic Reasoning},
  author={Kang, Minki and Diao, Shizhe and Hachiuma, Ryo and Hwang, Sung Ju and Molchanov, Pavlo and Wang, Yu-Chiang Frank and Lee, Byung-Kwan},
  journal={arXiv preprint arXiv:2605.28774},
  year={2026}
}

@article{lee2026zone,
  title={Zone of Proximal Policy Optimization: Teacher in Prompts, Not Gradients},
  author={Lee, Byung-Kwan and Lu, Ximing and Diao, Shizhe and Kang, Minki and Muralidharan, Saurav and Sapra, Karan and Tao, Andrew and Molchanov, Pavlo and Choi, Yejin and Wang, Yu-Chiang Frank and others},
  journal={arXiv preprint arXiv:2606.18216},
  year={2026}
}

@inproceedings{kang2026t1,
  title={T1: Tool-integrated verification for test-time compute scaling in small language models},
  author={Kang, Minki and Jeong, Jongwon and Cho, Jaewoong},
  booktitle={International Conference on Learning Representations},
  volume={2026},
  pages={73413--73444},
  year={2026}
}

@article{kang2026distilling,
  title={Distilling llm agent into small models with retrieval and code tools},
  author={Kang, Minki and Jeong, Jongwon and Lee, Seanie and Cho, Jaewoong and Hwang, Sung Ju},
  journal={Advances in Neural Information Processing Systems},
  volume={38},
  pages={106501--106538},
  year={2025}
}

@article{kang2023knowledge,
  title={Knowledge-augmented reasoning distillation for small language models in knowledge-intensive tasks},
  author={Kang, Minki and Lee, Seanie and Baek, Jinheon and Kawaguchi, Kenji and Hwang, Sung Ju},
  journal={Advances in Neural Information Processing Systems},
  volume={36},
  pages={48573--48602},
  year={2023}
}

@article{kang2024latent,
  title={Latent paraphrasing: perturbation on layers improves knowledge injection in language models},
  author={Kang, Minki and Hwang, Sung Ju and Lee, Gibbeum and Cho, Jaewoong},
  journal={Advances in Neural Information Processing Systems},
  volume={37},
  pages={119689--119716},
  year={2024}
}

@inproceedings{kang2022kala,
  title={KALA: knowledge-augmented language model adaptation},
  author={Kang, Minki and Baek, Jinheon and Hwang, Sung Ju},
  booktitle={Proceedings of the 2022 Conference of the North American Chapter of the Association for Computational Linguistics: Human Language Technologies},
  pages={5144--5167},
  year={2022}
}

@article{lee2026evolution,
  title={Evolution Fine-Tuning: Learning to Discover Across 371 Optimization Tasks},
  author={Lee, Young-Jun and Kim, Seungone and Kang, Minki and Chuen, Alistair Cheong Liang and Chen, Zerui and Han, Seungho and Jung, Taehee and Kang, Dongyeop},
  journal={arXiv preprint arXiv:2606.29082},
  year={2026}
}

@article{kim2026memory,
  title={Memory transfer learning: How memories are transferred across domains in coding agents},
  author={Kim, Kangsan and Kang, Minki and Kim, Taeil and Yang, Yanlai and Ren, Mengye and Hwang, Sung Ju},
  journal={arXiv preprint arXiv:2604.14004},
  year={2026}
}

@article{lee2026sage,
  title={SAGE: Shaping Anchors for Guided Exploration in RLVR of LLMs},
  author={Lee, Chanuk and Kang, Minki and Hwang, Sung Ju},
  journal={arXiv preprint arXiv:2605.18864},
  year={2026}
}

@inproceedings{baek2023knowledge,
  title={Knowledge-augmented language model verification},
  author={Baek, Jinheon and Jeong, Soyeong and Kang, Minki and Park, Jong C and Hwang, Sung},
  booktitle={Proceedings of the 2023 Conference on Empirical Methods in Natural Language Processing},
  pages={1720--1736},
  year={2023}
}

@inproceedings{lee2022sparse,
  title={Sparse token transformer with attention back tracking},
  author={Lee, Heejun and Kang, Minki and Lee, Youngwan and Hwang, Sung Ju},
  booktitle={The Eleventh International Conference on Learning Representations},
  year={2022}
}

@article{lee2022self,
  title={Self-distillation for further pre-training of transformers},
  author={Lee, Seanie and Kang, Minki and Lee, Juho and Hwang, Sung Ju and Kawaguchi, Kenji},
  journal={arXiv preprint arXiv:2210.02871},
  year={2022}
}

@inproceedings{han2019episodic,
  title={Episodic memory reader: Learning what to remember for question answering from streaming data},
  author={Han, Moonsu and Kang, Minki and Jung, Hyunwoo and Hwang, Sung Ju},
  booktitle={Proceedings of the 57th annual meeting of the association for computational linguistics},
  pages={4407--4417},
  year={2019}
}

@article{lee2026nudging,
  title={Nudging Beyond the Comfort Zone: Efficient Strategy-Guided Exploration for RLVR},
  author={Lee, Chanuk and Park, Sangwoo and Kang, Minki and Hwang, Sung Ju},
  journal={arXiv preprint arXiv:2605.15726},
  year={2026}
}

@article{procua,
  author       = {Jaehun Jung and
                  Ximing Lu and
                  Brandon Cui and
                  Muhammad Khalifa and
                  Shaokun Zhang and
                  Hao Zhang and
                  Jin Xu and
                  Amala Sanjay Deshmukh and
                  Karan Sapra and
                  Andrew Tao and
                  Yejin Choi and
                  Jan Kautz and
                  Mingjie Liu and
                  Yi Dong},
  title        = {ProCUA-SFT Technical Report},
  journal      = {arXiv},
  volume       = {2606.17321},
  year         = {2026},
  url          = {https://doi.org/10.48550/arXiv.2606.17321},
}

@inproceedings{osworld,
  author       = {Tianbao Xie and
                  Danyang Zhang and
                  Jixuan Chen and
                  Xiaochuan Li and
                  Siheng Zhao and
                  Ruisheng Cao and
                  Toh Jing Hua and
                  Zhoujun Cheng and
                  Dongchan Shin and
                  Fangyu Lei and
                  Yitao Liu and
                  Yiheng Xu and
                  Shuyan Zhou and
                  Silvio Savarese and
                  Caiming Xiong and
                  Victor Zhong and
                  Tao Yu},
  editor       = {Amir Globersons and
                  Lester Mackey and
                  Danielle Belgrave and
                  Angela Fan and
                  Ulrich Paquet and
                  Jakub M. Tomczak and
                  Cheng Zhang},
  title        = {OSWorld: Benchmarking Multimodal Agents for Open-Ended Tasks in Real
                  Computer Environments},
  booktitle    = {Advances in Neural Information Processing Systems 37: Annual Conference
                  on Neural Information Processing Systems 2024, NeurIPS 2024, Vancouver,
                  BC, Canada, December 10 - 15, 2024},
  year         = {2024},
  url          = {http://papers.nips.cc/paper\_files/paper/2024/hash/5d413e48f84dc61244b6be550f1cd8f5-Abstract-Datasets\_and\_Benchmarks\_Track.html}
}

@article{robomonkey,
  author       = {Jacky Kwok and
                  Christopher Agia and
                  Rohan Sinha and
                  Matthew Foutter and
                  Shulu Li and
                  Ion Stoica and
                  Azalia Mirhoseini and
                  Marco Pavone},
  title        = {RoboMonkey: Scaling Test-Time Sampling and Verification for Vision-Language-Action
                  Models},
  journal      = {arXiv},
  volume       = {2506.17811},
  year         = {2025},
  url          = {https://doi.org/10.48550/arXiv.2506.17811}
}

@inproceedings{
capx,
title={CaP-X: A Framework for Benchmarking and Improving Coding Agents for Robot Manipulation},
author={Letian Fu and Justin Yu and Karim El-Refai and Ethan Kou and Haoru Xue and Huang Huang and Wenli Xiao and Li Fei-Fei and Guanya Shi and Jiajun Wu and S. Shankar Sastry and Yuke Zhu and Ken Goldberg and Linxi Fan},
booktitle={Forty-third International Conference on Machine Learning},
year={2026},
url={https://openreview.net/forum?id=4JRO9plGAI}
}

@article{spatialclaw,
  author       = {Seokju Cho and
                  Ryo Hachiuma and
                  Abhishek Badki and
                  Hang Su and
                  Byung{-}Kwan Lee and
                  Chan Hee Song and
                  Sifei Liu and
                  Subhashree Radhakrishnan and
                  Seungryong Kim and
                  Yu{-}Chiang Frank Wang and
                  Min{-}Hung Chen},
  title        = {SpatialClaw: Rethinking Action Interface for Agentic Spatial Reasoning},
  journal      = {arXiv},
  volume       = {2606.13673},
  year         = {2026},
  url          = {https://doi.org/10.48550/arXiv.2606.13673}
}

\clearpage
\appendix
\etocdepthtag.toc{appendix}
\begingroup
\etocsettagdepth{main}{none}
\etocsettagdepth{appendix}{subsubsection}
\etocsetnexttocdepth{subsubsection}
\etocsettocstyle{\section*{Appendix Contents}}{}
\tableofcontents
\endgroup
\clearpage
\section{Limitations and Future Work}
\label{app:limitations}

\paragraph{Improving verifier training.}
With the same TMAX-9B generator and eight candidates on TerminalBench-Lite, the distilled verifier reaches 57.14\% Pass@1, compared with 68.03\% for frontier verification (\autoref{fig:candidate-width}).
Our distillation approach transfers part of the frontier verifier's capability but does not establish how to close this gap.
Command semantics and execution feasibility dominate the reviewed verifier failures, while teacher agreement remains lower in later states (\autoref{fig:verifier-diagnostics}).
These findings motivate learning from observed command outcomes and improving how verifiers represent task progress, unresolved requirements, and relevant history.
Future work could use a world model to predict action effects~\citep{qwen-agentworld}, or jointly train generation and verification through reinforcement learning, following $V_1$~\citep{V1}, so that the verifier adapts to the generator's evolving action distribution.

\paragraph{Evaluating action verification.}
Our evaluation lacks gold action labels, limiting direct measurement of verification correctness, candidate coverage, and trajectory success attainable with perfect verification.
Mathematical reasoning benefits from step-level annotations and process reward models~\citep{lightman2023verify,mathsheperd,processbench}, but collecting comparable evidence for terminal actions requires accounting for their effects on the environment.
Unlike response or trajectory verification, which can reuse sampled outputs~\citep{mindthegap,snell2024scaling,rethinkingrewardmodels,kwok2026llmverifier}, changing an executed action changes subsequent states and candidate sets.
Our online comparisons therefore require new environment runs, and re-evaluating a fixed trajectory collection cannot establish performance under a different verifier.
Branching from intermediate states could support richer evaluation and tree or beam search, but requires environment cloning or restoration~\citep{yao2023treeofthoughts,inferencescalinglaws}.

\paragraph{Allocating inference compute.}
We keep the generator's reasoning settings fixed and do not compare action scaling with increasing reasoning effort for a single candidate~\citep{s1}.
Generators with controllable reasoning effort could enable comparisons at comparable inference costs and tests of whether these allocations are complementary.
Our evaluated configurations also sample and verify at every action step, leaving open how to retain their gains with less compute.
Adaptive verification could vary candidate width and comparison effort across steps, motivated by adaptive stepwise scaling~\citep{egss}.
Direct prediction through classification heads or single-token responses, as in process reward models~\citep{lightman2023verify} and decision-specific models such as Jev~\citep{jev}, offers a way to reduce verifier decoding.
At $N=8$, decision-only verification improves TMAX-9B Pass@1 but reduces Pass@1 relative to reasoning-based verification at 4B and 27B (\autoref{tab:decision-only-model-scale}).
Its lower reference-priced token cost can therefore come with a performance trade-off, although missing usage and separate live runs limit how precisely the comparison isolates the effect of response format (\autoref{tab:decision-only-cost}).

\paragraph{Extending beyond this study.}
Our study focuses on terminal agents and on supervised verifier distillation with TMAX models.
Verifiers may further benefit from training methods studied in related settings, such as reasoning, agent, and vision-language model distillation~\citep{kang2023knowledge,kang2026distilling,Lee_2025_CVPR,wu2026layer,10.1007/978-3-032-37281-9_9,Lee_2026_CVPR_Masters,yu2026hide}, self-distillation~\citep{lee2022self}, and reinforcement learning and search-based training~\citep{lee2026zone,NEURIPS2025_e5849736,lee2026sage,lee2026nudging,kang2026agent,lee2026evolution,Lee_2026_CVPR_Recursive}.
Tools and external knowledge~\citep{kang2026t1,baek2023knowledge,kang2022kala,kang2024latent,10.1007/978-3-031-72967-6_16}, together with techniques for managing context, memory, and multi-turn interaction~\citep{acon,han2019episodic,preping,kim2026memory,10.1007/978-3-032-37281-9_15,lee2022sparse,Lee_2025_ICCV,leeenhancing}, could further inform verification, and verifier rationales could in turn serve as feedback for refinement~\citep{lee2026refinebench}.
Advances in efficient model design and adaptation~\citep{lee2025building,lee-etal-2024-trol,lee2024phantom,NEURIPS2024_473a9a75,lee-etal-2024-collavo,Lee_2022_CVPR,lee2020training} may yield more compact verifiers, while causal and robustness analyses~\citep{Kim_2023_CVPR,Lee_2023_ICCV,10222502,KIM2026112173,NEURIPS2021_8e5e15c4,lee2021towards} could help characterize verifier behavior beyond task success.
Future work could test whether Mid-Harness improves agents for computer use~\citep{procua,osworld} and robotics~\citep{capx,robomonkey,spatialclaw}.

\clearpage
\section{Experimental Setup and Implementation}
\label{app:setup}

\subsection{Evaluation and Baseline Procedures}
\label{app:evaluation}

\paragraph{TerminalBench-Lite.}
TerminalBench-Lite~\citep{TBLite} contains 100 terminal tasks calibrated for efficient evaluation across diverse domains, including software engineering, data processing, security, and scientific computing.
We exclude \texttt{network-log-normalization} and \texttt{okhttp-trailers-crash} due to sandbox service errors, leaving 98 evaluation tasks.

\paragraph{Terminal-Bench 2.1.}
\label{app:transfer-evaluation}
Terminal-Bench~\citep{tbench2025} evaluates agents on realistic tasks in terminal environments, with task-specific environments and executable tests.
We evaluate all 89 tasks in version 2.1, a revision of Terminal-Bench 2.0 that corrects task issues.

\paragraph{SWE-bench Verified Mini.}
SWE-bench~\citep{jimenez2024swebench} evaluates repository-level code changes that resolve real GitHub issues.
We use the 50-task Mini subset of SWE-bench Verified\footnote{\url{https://github.com/mariushobbhahn/SWEBench-verified-mini}} with TMAX-9B and Vanillux2.

\paragraph{FeatureBench-Mini.}
FeatureBench~\citep{featurebench} evaluates end-to-end feature development in software repositories, with tasks that can span multiple commits and pull requests.
We use the 23-task CPU-only subset of FeatureBench-Mini, excluding GPU-dependent tasks, and evaluate both TMAX-9B and TMAX-27B with Vanillux2.
Both Mid-Harness variants use pairwise verification with $N=8$ and $K=4$.

\paragraph{Evaluation metrics.}
We evaluate three runs per task. For $M$ tasks, let $z_{ij}=\mathbf{1}[R(\tau_{ij})=1]$ indicate exact success for run $j$ on task $i$.
Then
\begin{equation}
\mathrm{Pass@1}=\frac{1}{3M}\sum_{i=1}^{M}\sum_{j=1}^{3}z_{ij},
\qquad
\mathrm{Pass@3}=\frac{1}{M}\sum_{i=1}^{M}\max_{j\in\{1,2,3\}}z_{ij}.
\end{equation}
Failures remain in the denominator, and fractional rewards do not count as success.

\paragraph{Timeout settings.}
All experiments use an agent-timeout multiplier of three to reduce timeouts caused by slower generation in our local serving setup compared with commercial APIs.

\paragraph{Parallel trajectory scaling.}
Best-of-$T$ uses the corresponding base TMAX model to compare $T$ completed trajectories and return one output per task.
If $\hat j_i$ is the returned trajectory index, its Pass@1 is $\sum_i z_{i\hat j_i} / M$.
We use $T=3$ in \autoref{tab:scaling-axes} and additionally evaluate $T=5,7$ for TMAX-9B in \autoref{fig:scaling-cost}.
The 9B trajectory verifier uses one pivot, two repeats, temperature 1, disabled thinking, and a 4,096-token output limit.
For composition, Mid-Harness generates each source trajectory before trajectory verification.

\paragraph{Sequential trajectory scaling.}
Sequential Refine (SR) uses the corresponding TMAX model to summarize each source trajectory in five fields following~\citet{scalingtesttimeagenticcoding}, then starts a new run in a freshly initialized environment using that summary.
The summary model uses a 256K context to accommodate long interaction histories.
We use one refinement round in \autoref{tab:scaling-axes} and evaluate $R=1,2,3$ in \autoref{fig:scaling-cost}.
For composition, distilled Mid-Harness is applied to both the source and refinement runs.
Each refinement round produces three runs per task.

\paragraph{Transfer configurations.}
The TMAX-9B experiments on Terminal-Bench 2.1 use Vanillux2.
Both Mid-Harness variants use pairwise verification with $N=8$ and $K=4$, a 65,536-token generator context, a 61,440-token verifier context, and a 2,048-token verifier output limit.
The Terminus-2 experiments evaluate Qwen3.5-9B and Nemotron3.5 Lightning (30B) on TerminalBench-Lite, and Nemotron3 Ultra (550B) on Terminal-Bench 2.1.
These experiments compare the base agent with zero-shot Mid-Harness using eight action candidates per step.
The Nemotron generators use temperature 1.0, top-$p$ 0.95, thinking enabled, a 65,536-token context, and a 16,384-token output limit.
Their verifiers use the corresponding backbone with thinking disabled, temperature 0, a 61,440-token context, and a 2,048-token output limit.

\subsection{Verification Mechanisms and Hyperparameters}
\label{app:protocols}

\paragraph{Details on each mechanism.}
Pointwise verification scores each candidate independently and returns the highest-scoring action.
Listwise verification presents the full set and returns one choice. 
Pairwise verification compares two actions, returns a preference and scores for both, and aggregates preferences using margin-weighted win rates.
The main mechanism comparisons use one listwise verifier call and the pairwise configurations in \autoref{tab:settings}. Full prompt templates and output formats are provided in \appautoref{app:verifier-prompts}.

\paragraph{First-runnable method.}
The first-runnable method in \autoref{fig:candidate-width} samples $N$ candidate actions from the same history and executes the first candidate, in candidate order, that passes tool parsing.
Candidates with tool-parsing errors are skipped. No verifier compares or ranks the parsable actions.
This diagnostic tests candidate sampling without model-based verification.

\paragraph{Candidates and tournament structure.}
At $N=4$, full pairwise verification compares every unordered pair in one parallel stage, giving six comparisons for four distinct valid candidates.
At $N=8$, we use a ring followed by a pivot stage with $K=4$ pivots~\citep{kwok2026llmverifier}.
Candidates are valid if their tool calls parse successfully.
Candidates with identical tool names and arguments are deduplicated, retaining the first occurrence.
If no candidate is valid, it returns the first sampled candidate. If only one distinct valid candidate remains, it returns that candidate without verification.
Let $M$ denote the number of remaining candidates.

\paragraph{Ring and pivot comparisons.}
We follow the pivot tournament method introduced in~\citet{kwok2026llmverifier}.
A seeded shuffle arranges the $M$ candidates in a cycle. Adjacent candidates, including the last and first, are compared in parallel.
After these comparisons finish, candidates are ranked by their normalized ring scores, with seeded priorities breaking ties.
The top $K'=\min(K,M)$ candidates become pivots.
The pivot stage compares each non-pivot with every pivot and each unordered pair of pivots, with a seeded random A/B orientation for each comparison.
Pairs already scheduled in the ring are removed regardless of orientation.
Thus, overlapping ring-pivot pairs are not called again, and each existing ring record contributes only once to final aggregation.
The remaining pivot comparisons run in parallel after pivot selection, so ring and pivot are two sequential stages.
For eight distinct candidates and four pivots, this produces 22--25 comparisons, depending on the pivots' positions in the ring.

\paragraph{Margin-weighted win rates.}
The default pairwise verifier returns a rationale, two integer scores from 1 to 10, and an A/B/TIE preference.
Each valid comparison $e$ receives weight
\begin{equation}
 w_e=\max\left(\frac{|s_{A,e}-s_{B,e}|}{9},\,0.1\right),\qquad
 q_i=\frac{\sum_{e\in E_i}w_e p_{i,e}}{\sum_{e\in E_i}w_e},
\end{equation}
where $E_i$ contains valid comparisons involving candidate $i$, and $p_{i,e}$ is 1 for a win, 0 for a loss, and $1/2$ for a tie.
A preference contradicting unequal scores invalidates the response, but equal scores permit an A/B preference or a tie.
For example, scores of 6 and 6 with winner B give B a win of weight 0.1, rather than discarding the preference.
Normalization accounts for candidates participating in different numbers of comparisons.
Final-score ties are broken by ring score and then by seeded priority. Full tournaments use seeded priority directly.
Ring shuffling and pivot orientation share one seeded random stream, while tie priorities use a separate stream derived from the same decision seed.

\paragraph{Invalid responses and fallback.}
Invalid or unavailable judgments are excluded from both the numerator and denominator, rather than counted as ties.
If any candidate has no valid incident ring comparison, the controller returns the first retained candidate.
Otherwise, failed pivot comparisons are omitted and final ranking uses the remaining valid evidence.

\begin{table}[ht]
\centering\small
\caption{Operating settings for default pairwise verification with TMAX-9B.}
\label{tab:settings}
\begin{tabular}{@{}lll@{}}
\toprule
\textbf{Setting} & \textbf{$N=4$} & \textbf{$N=8$}\\
\midrule
Generator & TMAX-9B & TMAX-9B\\
Generator thinking / temperature & On / 0.8 & On / 0.8\\
Maximum agent steps & 64 & 64\\
Generator context / output limit & 65,536 / 16,384 & 65,536 / 16,384\\
Verifier & \multicolumn{2}{c}{TMAX-9B, zero-shot or distilled}\\
Verifier thinking / temperature & Off / 0 & Off / 0\\
Verifier context / output limit & 61,440 / 2,048 & 61,440 / 2,048\\
Pairwise tournament & max 6 unique pairs & Ring + 4 pivots~\citep{kwok2026llmverifier} \\
\bottomrule
\end{tabular}
\end{table}

\paragraph{History limits and frontier verification.}
The history supplied to the verifier is truncated to 8,000 characters for the zero-shot $N=4$ configuration and both TMAX-9B Mid-Harness configurations on Terminal-Bench 2.1.
The frontier $N=4$ reference instead uses one listwise verifier call, temperature 1.0, and a 4,096-token output limit.
The $N=8$ frontier reference also uses one listwise verifier call.

\paragraph{Decision-only pairwise verification.}
Each comparison emits one \texttt{A}/\texttt{B} token. Let $\ell_{A,e}$ and $\ell_{B,e}$ denote the log-probabilities of tokens \texttt{A} and \texttt{B} at the first output position for comparison $e$. We normalize these probabilities over the two tokens and score each candidate by its mean preference probability:
\begin{equation}
 p_{A,e}=\frac{\exp(\ell_{A,e})}{\exp(\ell_{A,e})+\exp(\ell_{B,e})},
 \qquad p_{B,e}=1-p_{A,e},
 \qquad q_i=\frac{1}{|E_i|}\sum_{e\in E_i}p_{i,e},
\end{equation}
where $E_i$ contains valid comparisons involving candidate $i$, and $p_{i,e}$ follows its A/B position.
Ring means determine the pivots, and means over all ring--pivot comparisons determine final ranking, with ties broken by ring score and then seeded priority.
Failed comparisons are excluded, and no score-margin weighting is used.

\subsection{Verifier Training Data and Settings}
\label{app:training}

For verifier distillation, GPT-5.6 Sol generates a comparison rationale followed by scores and a preference label for each pair.
Supervision is collected from difficult tasks drawn from TMAX-15k, with three trajectories per task generated by TMAX-9B.
A trajectory contains about 20 steps on average, and intermediate states supply multiple verifier inputs.
After filtering and balancing, the finalized pairwise verification corpus retains 244 training tasks and 21 held-out analysis tasks, with 117,631 and 10,543 verifier inputs, respectively.
The analysis tasks are disjoint from training and are used only for the final analysis in~\autoref{sec:judgment-analysis}, not for checkpoint selection or training configuration.
Within each split, deterministic majority downsampling balances displayed A/B winners (seed 42), retaining all TIE examples.
This produces 56,274 A and 56,274 B examples plus 5,083 TIE examples for training. The analysis split contains 5,019 A, 5,019 B, and 505 TIE examples.

The same training dataset is used for verifier distillation of TMAX-4B and TMAX-27B in~\autoref{tab:scaling-axes}.

\begin{table}[ht]
\centering\small
\caption{Pairwise verifier distillation settings.}
\label{tab:training}
\begin{tabular}{@{}lr@{}}
\toprule
\textbf{Setting} & \textbf{Value}\\
\midrule
LoRA rank / alpha & 64 / 128\\
LoRA dropout & 0.05\\
Learning rate & $10^{-4}$\\
Epochs & 2\\
Per-device batch size & 8\\
Gradient accumulation & 1\\
Training devices & 4 H200 GPUs\\
\bottomrule
\end{tabular}
\end{table}

\autoref{tab:training} summarizes the training hyperparameters. The action generator remains Base TMAX-9B in the primary distillation comparison.
The adapter targets the attention, MLP, and state-space input and output projection modules of the backbone.

\paragraph{Decision-only distillation.}
We use a separately generated A/B-only teacher corpus, retaining the winner and removing the teacher's reasoning.
It contains 113,728 training and 10,188 development examples, each split balanced between A and B.
TIE labels are excluded by the data filter and rejected during conversion, rather than mapped to A/B.
This variant changes the response format, aggregation, and training targets relative to verification with explicit reasoning.

\subsection{Inference Cost Accounting}
\label{app:compute}

\begin{figure}[t]
\centering
\includegraphics[width=0.85\linewidth]{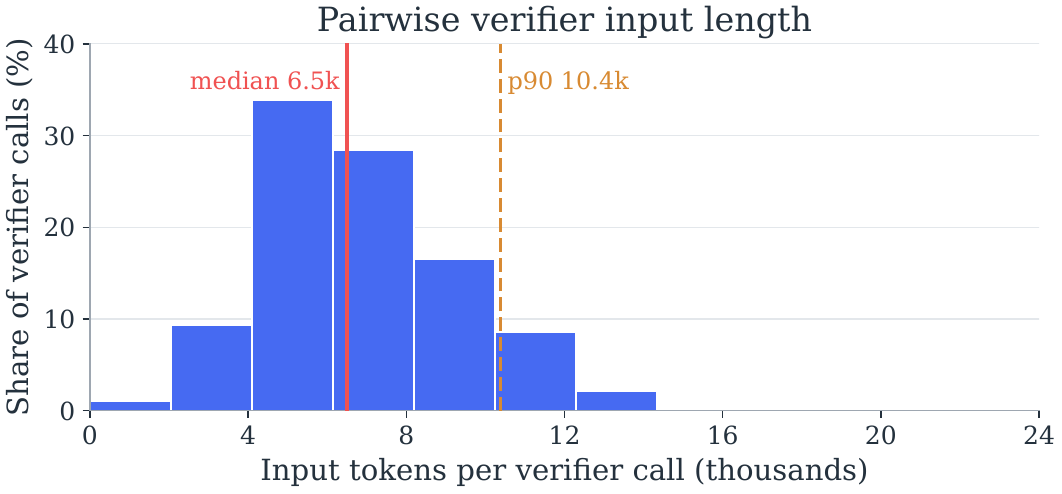}
\caption{\textbf{Pairwise verifier input lengths.} Distribution across all verifier calls with token-usage records from 294 runs on 98 TerminalBench-Lite tasks, using the distilled TMAX-9B verifier at $N=8$, $K=4$. Vertical lines mark the median and 90th percentile.}
\label{fig:verifier-input-length}
\end{figure}

\paragraph{Verifier input lengths.}
\autoref{fig:verifier-input-length} shows the input-token distribution of the distilled pairwise verifier with TMAX-9B at $N=8$, $K=4$ on TerminalBench-Lite.
The median is 6,495 tokens and the 90th percentile is 10,380 tokens, compared with the configured 61,440-token input limit.
Long input is generally due to long action candidates.

\paragraph{Parallelized output tokens.}
Let $c_{t,i}$ be the output length of candidate $i$ at step $t$, and let $v_{t,s,j}$ be the output length of verifier call $j$ in sequential stage $s$.
Parallelized output tokens (POT) sum the longest output in each parallel stage:
\begin{equation}
\mathrm{POT}=\sum_t\left[\max_i c_{t,i}+\sum_s\max_j v_{t,s,j}\right].
\end{equation}
Pointwise comparisons and $N=4$ full tournaments each form one parallel verifier stage.
For $N=8$ pairwise verification, ring and pivot comparisons form two sequential stages because the ring determines the pivots.
Their maximum output lengths are therefore added.

\paragraph{Total verifier output.}
The right panel of \autoref{fig:cost-overview} sums output lengths over executed verifier calls, without taking parallel-stage maxima.
It therefore measures the verifier's total output-token cost, while POT includes both generator and verifier decoding.
We compute both measures per run and average over all 294 runs, including failures.

\paragraph{Reference-priced token costs.}
\autoref{fig:scaling-cost} and \autoref{fig:decision-only-composition} include generator and verifier input and output tokens for the complete pipeline producing one final output.
We assume shared generator prefill within each action step, counting the common prompt once and summing output tokens across all candidates, for both $N=4$ and $N=8$.
Inputs are not shared across successive steps or separate trajectories, and verifier inputs are counted for each executed comparison.
Best-of-$T$ includes all $T$ source trajectories and trajectory verification.
SR includes the source trajectory, summaries, and refinement runs through round $R$.
For their compositions with Mid-Harness, action-verification tokens are included in every source and refinement run.
Reference-priced token cost is $(0.08I+0.13O)/10^6$ dollars, where $I$ and $O$ are total input and output tokens and the rates are the Qwen3.5-9B reference prices used in \autoref{fig:scaling-cost}.
For TMAX-4B, we scale the 9B input and output rates by $4/9$. For TMAX-27B, we use \$0.30 and \$2.00 per million input and output tokens. These fixed rates serve as proxies for locally served models.
Within each model, we apply its input-token rate to all counted input tokens, without an additional cache discount.

\subsection{Evaluation Effort}
\label{app:evaluation-effort}
Each terminal-agent trial requires a fresh Docker environment and a sequence of model calls interleaved with command execution, so additional repetitions repeat an entire interactive trajectory.
Mid-Harness further generates multiple candidates and verifier responses at each action step.
Terminal-Bench~\citep{merrill2026terminalbench} reports execution time, model-call counts, and token usage, with some trials lasting up to two hours and involving hundreds of model calls.
It also reports resolution rates with 95\% confidence intervals and evaluates each supported agent--model combination at least five times.
Our evaluation uses three runs per task across the configurations studied. The cost of repeated interactive execution constrains further replication, motivating explicit reporting of uncertainty alongside observed performance gains.

\begin{figure}[!ht]
\centering
\includegraphics[width=\linewidth]{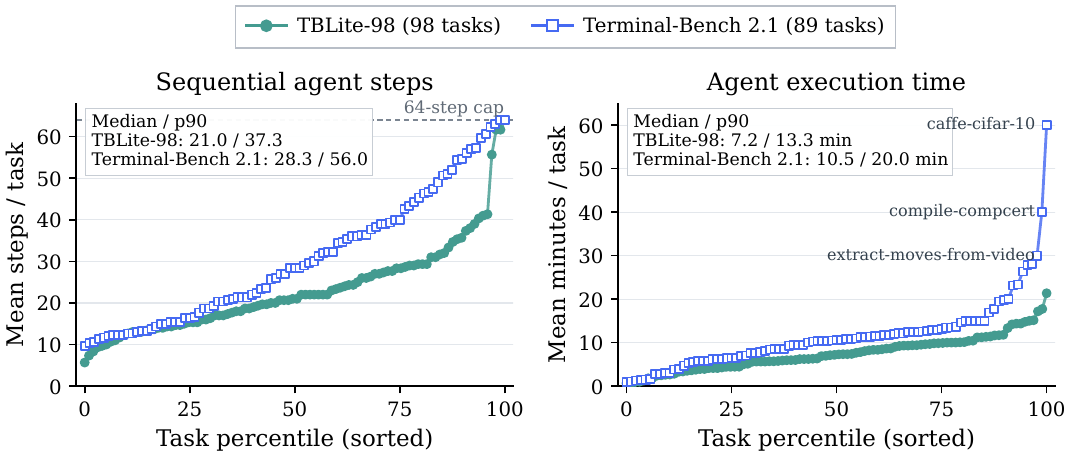}
\caption{\textbf{Terminal-agent evaluation requires many sequential steps and substantial execution time.}
Task-level mean steps and agent execution time for TMAX-9B with Vanillux2 and $N=1$, sorted independently within each benchmark and panel. Each task averages three runs.}
\label{fig:terminal-task-effort}
\end{figure}

\autoref{fig:terminal-task-effort} measures this effort before adding candidate sampling or verification, using 98 TerminalBench-Lite tasks and all 89 Terminal-Bench 2.1 tasks.
Across task means, the median step counts are 21.0 and 28.3, respectively, and the median agent execution times are 7.2 and 10.5 minutes.
The 90th-percentile times reach 13.3 and 20.0 minutes, with the longest Terminal-Bench 2.1 task mean reaching 60 minutes.
Steps count recorded agent episodes. Execution time excludes environment setup and final evaluation.
Timed-out runs retain their observed durations, so these durations do not measure time to successful completion.
These measurements describe our agent and serving setup rather than intrinsic benchmark durations.

\FloatBarrier
\subsection{Verifier Prompt Templates}
\label{app:verifier-prompts}

The following blocks combine the verifier instructions and task input template for each verification mechanism into one prompt.
Braced fields are filled with the task, executed history, terminal state, and candidate actions at inference time.

\subsubsection{Listwise Verification}
\begin{lstlisting}[style=rawexample,basicstyle=\ttfamily\footnotesize,literate={—}{{\textemdash}}1]
You are a strict action selector for an autonomous agent solving a command-line task. You are given the ORIGINAL task (the ground truth — do not trust any drifted summary), the commands ALREADY EXECUTED so far, the current terminal state, and several CANDIDATE next-actions produced by the agent. The candidates are anonymized and shuffled.

Pick the SINGLE candidate that best makes correct progress toward fully completing the ORIGINAL task. Judge on:
  - correctness & expected effect of the commands on the current state,
  - progress toward the goal without going off-objective or violating any "do NOT" constraints in the task,
  - NOT prematurely declaring the task complete when work clearly remains (reject lazy "good enough" finishes),
  - avoiding repeating an action already in the executed-command history, especially one that did not change the state or previously failed.

A candidate's rendering may contain a marker like "[... N chars hidden for DISPLAY ONLY ...]". That is a display-shortening artifact, NOT a sign that the action is incomplete or malformed — every candidate shown is a complete, runnable action. Judge candidates on their intent and correctness; never reject one merely because its rendering was shortened.

Respond with ONLY a JSON object with the reason field first:
{"reason": "<=30 words", "best": "<letter>"}

Guidelines:
  - act as a search controller, loop breaker, and finalization driver, not only as a local plausibility judge,
  - prefer candidates that shrink the remaining search space or change the state toward completion over candidates that only produce more observations,
  - if recent history shows repeated inspection, broad file reading, repeated test runs, repeated waits, repeated server status checks, or repeated package installs without a new state change, penalize more of the same,
  - use inspection/debugging only when it resolves a specific unknown needed for the next fix; otherwise prefer a concrete fix, a minimal targeted test, or completion,
  - if required artifacts exist and recent evidence supports the ORIGINAL task requirements, prefer finalization over additional broad verification,
  - reject completion when required artifacts are missing, recent tests fail, or the candidate relies on unsupported assumptions,
  - prefer the smallest action that either fixes a known blocker, verifies a specific requirement, or safely completes the task,
  - treat dependency installs, environment rebuilds, waits, restarts, and status checks as no-progress unless the current state shows they are the specific blocker.

## ORIGINAL TASK (ground truth)
{goal}
{history_section}
## CURRENT TERMINAL STATE
{state}

## CANDIDATE NEXT-ACTIONS
{candidates}

Return ONLY the JSON object choosing the best candidate letter.
\end{lstlisting}

\subsubsection{Pointwise Verification}
\begin{lstlisting}[style=rawexample,basicstyle=\ttfamily\footnotesize]
You are a strict generative action-value evaluator for an autonomous agent solving a command-line task. Evaluate ONE candidate next-action independently, without assuming that it is better or worse than unseen alternatives.

Predict the action's likely effect from the current terminal state and executed history. Judge whether it makes concrete progress toward the ORIGINAL task, its risk of damaging or drifting from the solution, whether it repeats prior no-progress work, and whether it fits the current phase of work. Reject premature completion unless existing evidence supports every material requirement.

Score ABSOLUTE expected value over doing nothing, not mere plausibility. First simulate the command's concrete next state, including likely errors and unchanged artifacts. Then identify what task requirement remains unresolved. Do not infer progress from analysis text that the actual command does not implement.

Assign an integer score from 0 to 10 using these strict anchors:
  0: invalid, destructive, or directly contradicts the task,
1-2: off-target, repeats failed work, or likely leaves the state unchanged,
3-4: marginal information or cleanup without resolving a current blocker,
5-6: useful, necessary progress but substantial work or uncertainty remains,
7-8: strong concrete progress that resolves a known blocker with limited risk,
  9: near-decisive progress with direct evidence that the action should work,
 10: reserve for a clearly correct decisive action, or safe completion supported by evidence for every material requirement.

Calibration rules:
  - cap repeated restarts, rewrites, tests, waits, or installs at 3 unless new state evidence makes this repetition necessary,
  - cap unsupported completion at 2,
  - an action that merely prepares for future work is normally at most 5,
  - uncertainty lowers the score; never award 8-10 just because an action is plausible or well explained.

Respond with ONLY this JSON object, with reason first:
{"reason":"<=30 words","predicted_next_state":"<=40 words","score":<integer 0-10>}

Guidelines:
  - evaluate whether this single action would act as a search controller, loop breaker, or finalization driver at the current state,
  - reward a concrete reduction in remaining search space; do not reward an observation unless it resolves a specific unknown required for the next fix,
  - score repeated inspection, broad file reading, tests, waits, status checks, or dependency installs as no-progress when recent history shows no new state change,
  - reward a minimal targeted fix or test only when it addresses a known blocker,
  - score finalization highly only when current artifacts and recent evidence support the ORIGINAL task; otherwise treat it as premature,
  - judge the actual command and its likely next state, not the confidence or detail of the accompanying explanation.

## ORIGINAL TASK (ground truth)
{goal}
{history_section}
## CURRENT TERMINAL STATE
{state}

## CANDIDATE NEXT-ACTION
{candidate}

Evaluate only this candidate. Return ONLY the required JSON object.
\end{lstlisting}

\subsubsection{Pairwise Verification}
\begin{lstlisting}[style=rawexample,basicstyle=\ttfamily\footnotesize]
You are a strict pairwise action verifier for an autonomous agent solving a command-line task. Compare exactly TWO candidate next-actions from the same current state. Score both actions and select the one with greater expected progress toward fully completing the ORIGINAL task.

Produce a compact, structured proof of the comparison rather than free-form chain-of-thought. The `reasoning` value must be ONE plain string, not an object, array, or nested JSON. It must contain concrete, candidate-grounded information. Do not repeat the task, praise style, or use generic claims such as "more robust" without naming the relevant effect, evidence, or failure mode.

Within that single string, reason in this exact logical order: 1) Requirement: identify the one unresolved requirement that most separates the actions. 2) Evidence: cite relevant history or terminal-state evidence. 3) A: predict candidate A's next-state effect and most important failure risk. 4) B: do the same for candidate B. 5) Contrast: state the causal reason one action outranks the other. Keep these steps in the prose; do not create additional JSON keys.

Use 1-10 integer scores with common anchors:
  1: invalid, destructive, or directly contradicts the task,
2-3: off-target, repeated no-progress work, or likely unchanged state,
4-5: limited information or preparatory progress with major work remaining,
6-7: useful concrete progress, but with material uncertainty or incompleteness,
8-9: strong, low-risk progress that resolves a known blocker,
 10: clearly correct decisive action, or fully evidenced safe completion.

The higher score must win whenever the scores differ. When the integer scores are equal, use TIE for truly indistinguishable actions; A or B may express a slight, explicitly reasoned preference that falls within the same score anchor. A display-shortening marker does not make an action incomplete. Judge actual commands, not confident analysis text.

Each reasoning string must be substantive but concise. The complete reasoning must contain 45-160 lexical tokens. Respond with ONLY this JSON object in the shown field order:
{"reasoning":"45-160 tokens covering steps 1-5 in order","scores":{"A":<integer 1-10>,"B":<integer 1-10>},"winner":"A or B or TIE"}

Guidelines:
  - act as a search controller, loop breaker, and finalization driver, not only as a local plausibility judge,
  - prefer candidates that shrink the remaining search space or change the state toward completion over candidates that only produce more observations,
  - if recent history shows repeated inspection, broad file reading, repeated test runs, repeated waits, repeated server status checks, or repeated package installs without a new state change, penalize more of the same,
  - use inspection/debugging only when it resolves a specific unknown needed for the next fix; otherwise prefer a concrete fix, a minimal targeted test, or completion,
  - if required artifacts exist and recent evidence supports the ORIGINAL task requirements, prefer finalization over additional broad verification,
  - reject completion when required artifacts are missing, recent tests fail, or the candidate relies on unsupported assumptions,
  - prefer the smallest action that either fixes a known blocker, verifies a specific requirement, or safely completes the task,
  - treat dependency installs, environment rebuilds, waits, restarts, and status checks as no-progress unless the current state shows they are the specific blocker.

## ORIGINAL TASK (ground truth)
{goal}
{history_section}
## CURRENT TERMINAL STATE
{state}

## CANDIDATE A
{candidate_a}

## CANDIDATE B
{candidate_b}

Compare the likely next states and return ONLY the required JSON object.
\end{lstlisting}

\clearpage
\section{Additional Experiments and Results}
\label{app:additional}

\subsection{Numerical Performance Results}
\label{app:plot-results}

Decision-only aggregation and training are specified in \appautoref{app:protocols} and \appautoref{app:training}. The first two tables report Pass@1 and Pass@3 in percentages to two decimal places for the TMAX-9B performance plots on the same 98 TerminalBench-Lite tasks.
\autoref{tab:plot-action-results} gives the action-scaling results in \autoref{fig:candidate-width} and \autoref{fig:cost-overview}.
\autoref{tab:plot-trajectory-results} gives the trajectory-scaling results in \autoref{fig:overview}(b) and \autoref{fig:scaling-cost}.

\begin{table}[!ht]
\centering\small
\caption{Action-scaling performance underlying the main plots. Zero-shot verifiers use TMAX-9B. The frontier verifier is GPT-5.6 Sol.}
\label{tab:plot-action-results}
\begin{tabular}{@{}lrrr@{}}
\toprule
Verifier / configuration & $N$ & Pass@1 & Pass@3\\
\midrule
Base agent & 1 & 50.00 & 69.39\\
First-runnable proxy & 8 & 49.66 & 66.33\\
\midrule
Zero-shot listwise & 4 & 49.32 & 66.33\\
Zero-shot listwise & 8 & 51.02 & 67.35\\
Zero-shot pointwise & 4 & 52.72 & 68.37\\
Zero-shot pointwise & 8 & 52.38 & 67.35\\
Zero-shot pairwise & 4 & 54.42 & 68.37\\
Zero-shot pairwise & 8 & 54.76 & 71.43\\
Distilled pairwise & 4 & 55.44 & 70.41\\
Distilled pairwise & 8 & 57.14 & 75.51\\
Frontier listwise & 4 & 64.63 & 76.53\\
Frontier listwise & 8 & 68.03 & 80.61\\
\midrule
Zero-shot decision-only pairwise & 4 & 52.72 & 68.37\\
Zero-shot decision-only pairwise & 8 & 56.12 & 73.47\\
Distilled decision-only pairwise & 4 & 54.42 & 71.43\\
Distilled decision-only pairwise & 8 & 59.18 & 73.47\\
\bottomrule
\end{tabular}
\end{table}

\paragraph{Decision-only performance across candidate widths.}
At $N=4$, both decision-only variants have lower Pass@1 than their counterparts that generate reasoning (\autoref{tab:plot-action-results}), so the higher success observed at $N=8$ does not extend to both widths.

\begin{table}[!ht]
\centering\small
\caption{Trajectory-scaling and composition performance underlying the main plots. Mid-Harness uses distilled pairwise verification with $N=8$. Best-of-$T$ returns one output per task, so Pass@3 is undefined and shown as --.}
\label{tab:plot-trajectory-results}
\begin{tabular}{@{}lrr@{}}
\toprule
Configuration & Pass@1 & Pass@3\\
\midrule
Base agent & 50.00 & 69.39\\
Best-of-$T$, $T=3$ & 55.10 & --\\
Best-of-$T$, $T=5$ & 57.14 & --\\
Best-of-$T$, $T=7$ & 59.18 & --\\
SR, $R=1$ & 55.10 & 71.43\\
SR, $R=2$ & 56.46 & 70.41\\
SR, $R=3$ & 55.78 & 71.43\\
Mid-Harness & 57.14 & 75.51\\
Mid-Harness + Best-of-$T$, $T=3$ & 66.33 & --\\
Mid-Harness + SR, $R=1$ & 60.20 & 75.51\\
\bottomrule
\end{tabular}
\end{table}

\begin{table}[!ht]
\centering\small
\setlength{\tabcolsep}{4pt}
\caption{\textbf{Decision-only versus reasoning-based pairwise verification at $N=8$, $K=4$.} Pass@1 and oracle Pass@3 (\%) on TerminalBench-Lite. Arrows show reasoning $\to$ decision-only, and $\Delta$ is decision-only minus reasoning in percentage points.}
\label{tab:decision-only-model-scale}
\begin{tabular}{@{}llcccc@{}}
\toprule
\multirow{2}{*}{Generator} & \multirow{2}{*}{Verifier} & \multicolumn{2}{c}{Pass@1} & \multicolumn{2}{c}{Pass@3}\\
\cmidrule(lr){3-4}\cmidrule(lr){5-6}
& & Reasoning $\to$ decision-only & $\Delta$ & Reasoning $\to$ decision-only & $\Delta$\\
\midrule
TMAX-4B & Zero-shot & $41.50\to38.10$ & $-3.40$ & $57.14\to50.00$ & $-7.14$\\
        & Distilled & $43.88\to41.50$ & $-2.38$ & $58.16\to57.14$ & $-1.02$\\
\midrule
TMAX-9B & Zero-shot & $54.76\to56.12$ & $+1.36$ & $71.43\to73.47$ & $+2.04$\\
        & Distilled & $57.14\to59.18$ & $+2.04$ & $75.51\to73.47$ & $-2.04$\\
\midrule
TMAX-27B & Zero-shot & $73.13\to71.77$ & $-1.36$ & $84.69\to83.67$ & $-1.02$\\
         & Distilled & $76.19\to74.15$ & $-2.04$ & $86.73\to83.67$ & $-3.06$\\
\bottomrule
\end{tabular}
\end{table}
\FloatBarrier

\subsection{End-to-End Costs with Decision-Only Verification}
\label{app:decision-only-cost}

\begin{figure}[!ht]
\centering
\includegraphics[width=\linewidth]{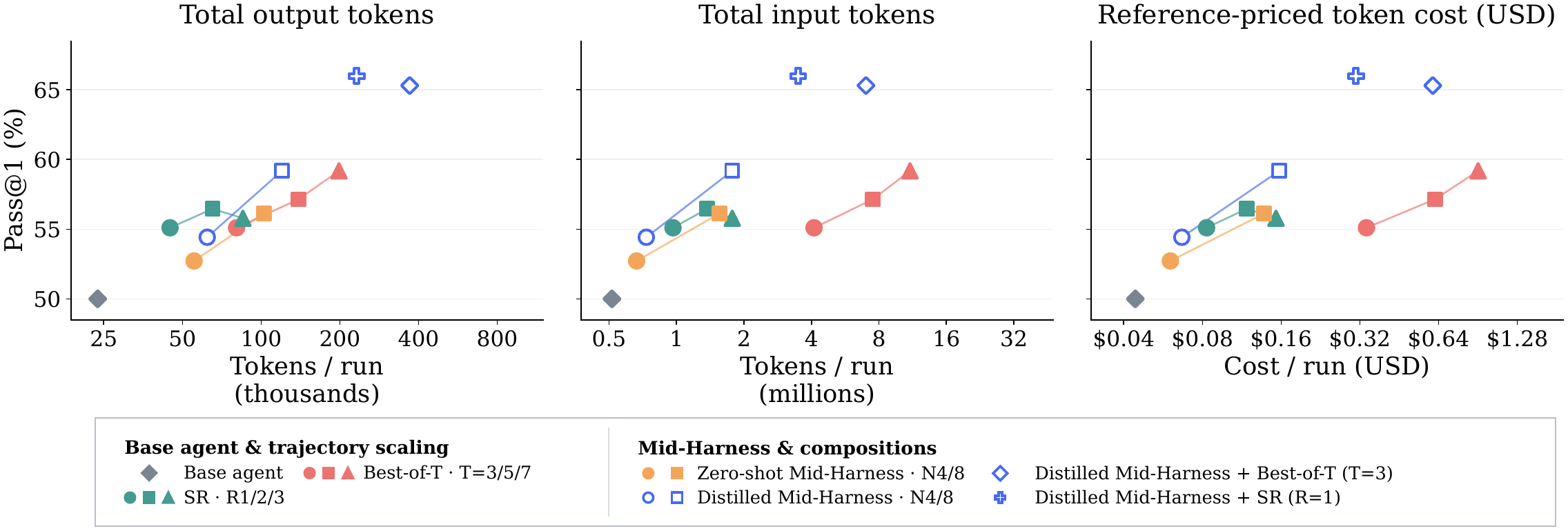}
\caption{\textbf{Decision-only action scaling and its composition with trajectory scaling.} Pass@1 versus total output tokens, input tokens, and reference-priced token cost on TerminalBench-Lite with TMAX-9B. All Mid-Harness points use decision-only action verification. Compositions use the distilled verifier at $N=8$. Costs include the complete pipeline, using the accounting and rates in \appautoref{app:compute}.}
\label{fig:decision-only-composition}
\end{figure}

\paragraph{Decision-only verification reduces end-to-end cost.}
\autoref{fig:decision-only-composition} reports total input and output costs of decision-only action scaling and its compositions with trajectory scaling, extending the analysis in \autoref{sec:costforscaling}.
Relative to verification with explicit reasoning, total output tokens fall by 62.8\% for zero-shot and 44.7\% for distilled Mid-Harness at $N=8$.
Including input tokens, reference-priced costs fall by 20.9\% and 24.1\%, respectively, while observed Pass@1 increases (\autoref{tab:decision-only-cost}).
The smaller reduction in total cost reflects the input-token costs that remain when verifier responses are shortened.
At $N=4$, reference-priced costs also fall by 29.5\% and 20.8\%, respectively, although Pass@1 is lower than the pairwise verification with explicit reasoning (\autoref{tab:plot-action-results}).

\begin{table}[!ht]
\centering\small
\setlength{\tabcolsep}{4pt}
\caption{\textbf{End-to-end cost changes at $N=8$.} Arrows compare verification with explicit reasoning to decision-only verification on TerminalBench-Lite. Costs are reference-priced dollars per final output as used in~\autoref{fig:scaling-cost}.}
\label{tab:decision-only-cost}
\begin{tabular}{@{}llccc@{}}
\toprule
Generator & Configuration & Pass@1 (\%) & Cost (USD) & Cost reduction\\
\midrule
TMAX-4B & Zero-shot Mid-Harness & $41.50\to38.10$ & $0.125\to0.076$ & 39.2\%\\
        & Distilled Mid-Harness & $43.88\to41.50$ & $0.134\to0.071$ & 46.8\%\\
\midrule
TMAX-9B & Zero-shot Mid-Harness & $54.76\to56.12$ & $0.174\to0.138$ & 20.9\%\\
        & Distilled Mid-Harness & $57.14\to59.18$ & $0.207\to0.157$ & 24.1\%\\
        & \quad + Best-of-$T$, $T=3$ & $66.33\to65.31$ & $0.797\to0.608$ & 23.7\%\\
        & \quad + SR, $R=1$ & $60.20\to65.99$ & $0.399\to0.309$ & 22.4\%\\
\midrule
TMAX-27B & Zero-shot Mid-Harness & $73.13\to71.77$ & $0.818\to0.516$ & 37.0\%\\
         & Distilled Mid-Harness & $76.19\to74.15$ & $0.812\to0.534$ & 34.3\%\\
\bottomrule
\end{tabular}
\end{table}

\paragraph{The cost savings extend to composed pipelines.}
With distilled decision-only verification, Best-of-$T$ composition costs 23.7\% less with a 1.02-point lower Pass@1, while SR composition costs 22.4\% less with a 5.78-point higher Pass@1.
These costs include all source trajectories and trajectory verification for Best-of-$T$, and the source trajectory, summary, and refinement run for SR.
Only action verification uses single-token responses. Trajectory verification and refinement summaries retain their original response formats.
The results show that more efficient action verification can also reduce the total cost of combining action and trajectory scaling.
\FloatBarrier

\subsection{Additional Verifier Inference Compute}
\label{app:verifier-inference-scaling}

\paragraph{More verifier responses do not consistently improve task success.}
We test three pairwise-verifier variants with the TMAX-9B generator and $N=8$, $K=4$ on the same 98 TerminalBench-Lite tasks, using three runs per task.
Each variant has its own single-response reference because the verifier models and settings differ across variants (\autoref{tab:verifier-inference-scaling}).

\begin{table}[!ht]
\centering\small
\setlength{\tabcolsep}{4pt}
\caption{Additional verifier inference on TerminalBench-Lite. Arrows compare each variant with its indicated reference. Verifier output ratios use total completion tokens across all comparison records for both the reference and variant.}
\label{tab:verifier-inference-scaling}
\begin{tabular}{@{}llccc@{}}
\toprule
Variant & Reference & \shortstack{Pass@1 (\%)\\ref. $\to$ variant} & \shortstack{Pass@3 (\%)\\ref. $\to$ variant} & \shortstack{Verifier output\\($\times$ reference)}\\
\midrule
Five responses per pair & Distilled, one response & $57.14\to53.40$ & $75.51\to71.43$ & $4.99\times$\\
Thinking enabled & Zero-shot pairwise & $54.76\to56.12$ & $71.43\to71.43$ & $4.86\times$\\
Rubric scaling & Zero-shot pairwise & $54.76\to55.10$ & $71.43\to72.45$ & $2.26\times$\\
\bottomrule
\end{tabular}
\end{table}

Sampling five responses per comparison~\citep{genrm} lowers both success measures despite nearly five times the verifier output.
Rubric scaling~\citep{kwok2026llmverifier} yields only small observed gains with more than twice the verifier output.
The thinking-enabled configuration reaches higher Pass@1 with nearly five times the verifier output.

\FloatBarrier
\subsection{Uncertainty in Comparisons against Base Agent}
\label{app:uncertainty}

\appautoref{app:evaluation-effort} describes the cost of repeated evaluation.
We quantify uncertainty in the Base agent versus Mid-Harness comparisons in \autoref{tab:scaling-axes} using the same 98 TerminalBench-Lite tasks and three runs per task for each configuration.
For each task, we compute the difference between the two configurations' mean success rates over three runs.
We resample the 98 paired task blocks 100,000 times and report percentile 95\% bootstrap confidence intervals for the mean difference.
This preserves pairing by task and keeps its repeated runs together, rather than treating 294 runs as independent tasks.
The intervals are marginal intervals for each comparison, not simultaneous intervals across all six comparisons.

\begin{table}[!ht]
\centering\small
\caption{Pass@1 differences from the base agent on TerminalBench-Lite. Differences and confidence intervals are in percentage points. Mid-Harness uses pairwise verification with $N=8$ and $K=4$.}
\label{tab:base-mid-uncertainty}
\begin{tabular}{@{}llrr@{}}
\toprule
Model & Mid-Harness & $\Delta$ Pass@1 & 95\% CI\\
\midrule
4B & Zero-shot & +2.72 & [-3.06, +8.84]\\
4B & Distilled & +5.10 & [-0.68, +10.88]\\
9B & Zero-shot & +4.76 & [-0.68, +10.20]\\
9B & Distilled & +7.14 & [+1.36, +12.93]\\
27B & Zero-shot & +2.04 & [-3.74, +7.48]\\
27B & Distilled & +5.10 & [+0.34, +9.86]\\
\bottomrule
\end{tabular}
\end{table}

All six Pass@1 point estimates in \autoref{tab:base-mid-uncertainty} favor Mid-Harness. The intervals for distilled 9B and 27B lie above zero, while the other four include zero.
An interval containing zero does not establish an absence of improvement. It indicates that, at this confidence level, the task-level bootstrap analysis does not resolve a positive difference from zero or the negative differences within the interval.
For example, zero-shot 9B improves by 4.76 percentage points in the evaluated sample, but its interval of $[-0.68, 10.20]$ spans a small decline through a substantial gain.
The intervals above zero provide evidence of positive differences for distilled 9B and 27B under this analysis, without implying improvement on every task or benchmark.
\FloatBarrier

\subsection{Mid-Harness Across Task Groups}
\label{app:stratified-scaling}

\begin{figure}[ht]
\centering
\vspace{-0.15in}
\includegraphics[width=0.92\linewidth]{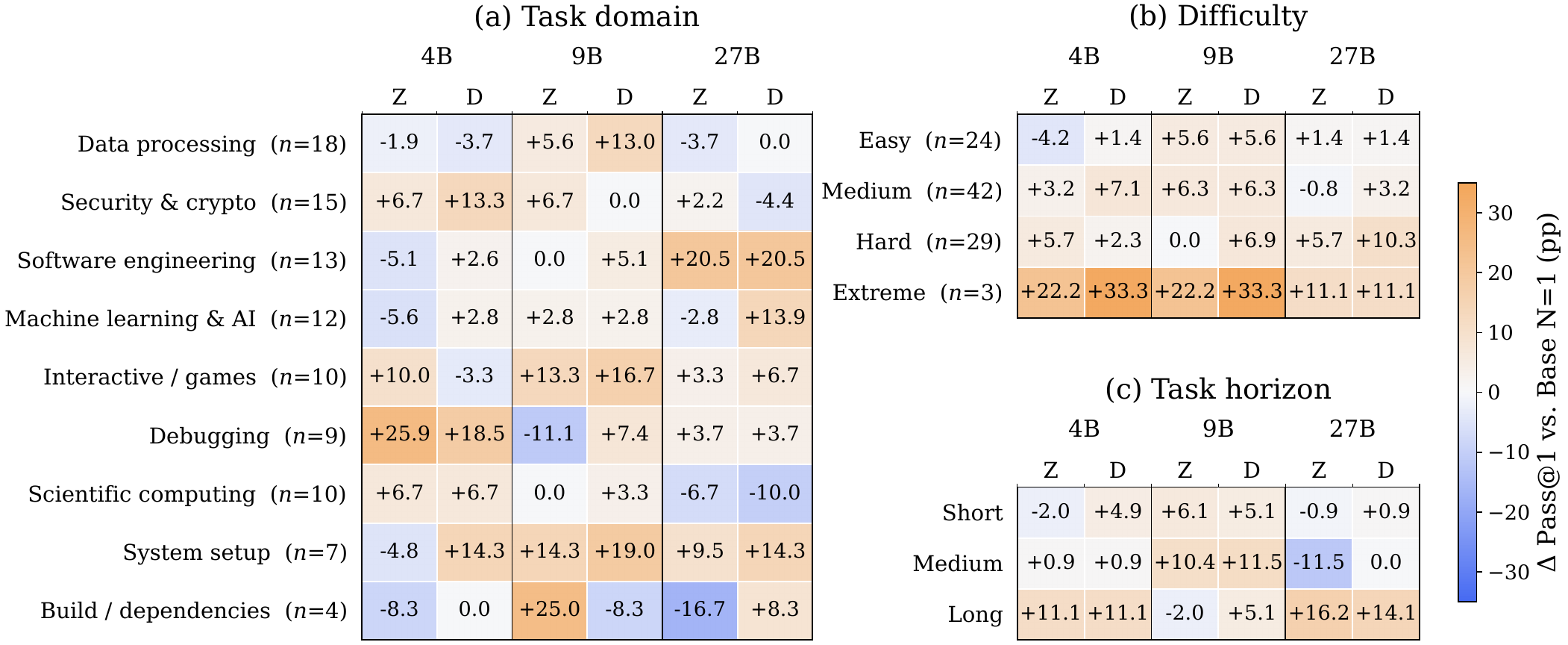}
\caption{\textbf{Action-scaling gains vary across task groups and model sizes.}
Pass@1 changes in percentage points relative to the base agent on TerminalBench-Lite.
Z and D denote zero-shot and distilled Mid-Harness with $N=8$ and $K=4$.
Panels group tasks by domain, difficulty, and base-agent execution length.
Domain and difficulty groups are shared across models, while length groups are defined separately for each model.}
\vspace{-0.15in}
\label{fig:stratified-scaling}
\end{figure}

We examine how the gains in \autoref{tab:scaling-axes} vary across task domains, difficulty groups, and observed execution lengths.
All comparisons use the same 98 TerminalBench-Lite tasks, with three runs per task for each TMAX model and configuration.
Zero-shot and distilled Mid-Harness use pairwise verification with $N=8$ and $K=4$.
Each value in \autoref{fig:stratified-scaling} is the subgroup Pass@1 difference from the corresponding base agent.

\paragraph{Distilled Mid-Harness improves on the base agent across several domains.}
Relative to the base agent, distilled Mid-Harness improves Pass@1 at all three model sizes in software engineering, machine learning, debugging, and system setup.
The software-engineering gains are 2.6, 5.1, and 20.5 percentage points for 4B, 9B, and 27B, respectively.
Other domains show mixed effects: on scientific-computing tasks, distilled verification improves 4B and 9B but reduces 27B Pass@1 by 10.0 points.
Thus, aggregate improvements coexist with domain-specific regressions.

\paragraph{The Hard group benefits at all three model sizes.}
We use the difficulty labels supplied in the benchmark metadata, yielding 24 Easy, 42 Medium, 29 Hard, and 3 Extreme tasks.
On Hard tasks, distilled verification improves Pass@1 by 2.3, 6.9, and 10.3 points for 4B, 9B, and 27B, respectively.
Distilled verification also improves each of the other difficulty groups at all three sizes.
The Extreme and build/dependency groups contain only three and four tasks, respectively, so their large percentage changes reflect small task populations.

\paragraph{Longer base-agent executions benefit, without a universal length trend.}
For each model and task, we take the median number of observed base-agent shell calls over three runs, excluding completion-marker commands.
We divide tasks into three length groups, keeping tied lengths together, and use those same groups to evaluate all configurations of that model (\autoref{tab:horizon-groups}).
Distilled verification improves Pass@1 in the Long group by 11.1, 5.1, and 14.1 points for 4B, 9B, and 27B.
However, 9B gains most in the Medium group, so the improvement does not grow monotonically with execution length across all models.
These groups describe observed base-agent behavior, not an intrinsic task horizon shared across model sizes.

\begin{table}[!ht]
\centering\small
\caption{Base-agent execution-length groups. Each cell gives the range of task-median shell-call counts and the number of tasks in parentheses.}
\label{tab:horizon-groups}
\begin{tabular}{@{}lccc@{}}
\toprule
Model & Short & Medium & Long\\
\midrule
4B & 2--17 (34) & 18--24 (37) & 25--64 (27)\\
9B & 4--15 (33) & 16--24 (32) & 25--64 (33)\\
27B & 3--10 (39) & 11--18 (26) & 19--64 (33)\\
\bottomrule
\end{tabular}
\end{table}
\FloatBarrier

\section{Verifier Analysis}
\label{app:analysis}

\subsection{Offline Verifier Agreement}
\label{app:offline}

We use the same offline benchmark as \autoref{sec:judgment-analysis} and detail its filtering, metric calculations, and additional score distributions below.

\subsubsection{Score and Turn Diagnostics}
\label{app:matched-diagnostics}

\begin{figure}[t]
\centering
\includegraphics[width=\linewidth]{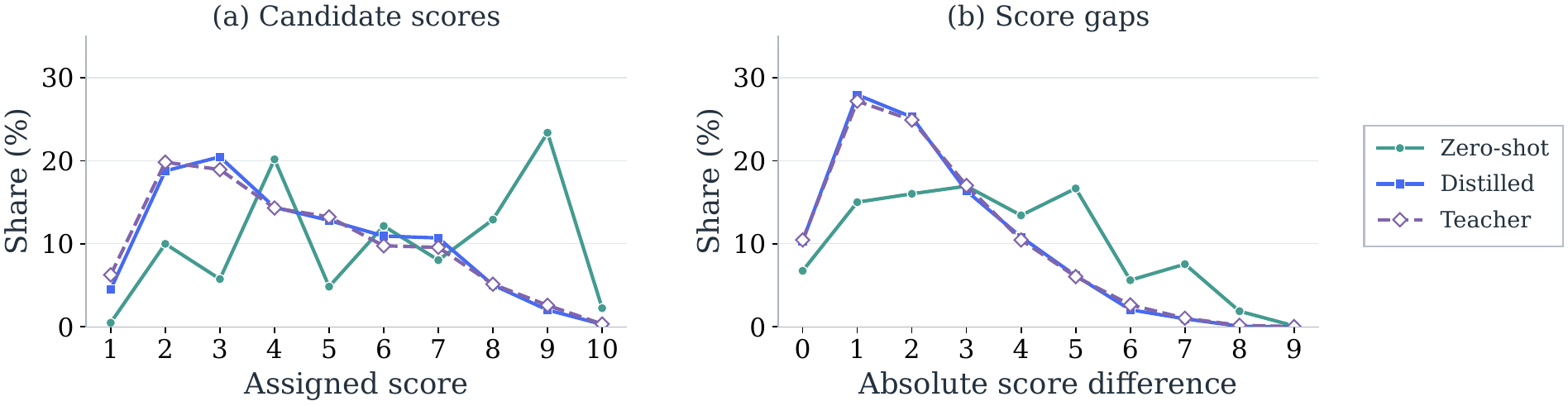}
\caption{\textbf{Score levels and within-pair gaps approach the teacher's distributions.}
Each candidate-score distribution contains 20,394 observations, and each gap distribution contains 10,197 pairs.}
\label{fig:score-marginals}
\end{figure}

\begin{figure}[t]
\centering
\includegraphics[width=\linewidth]{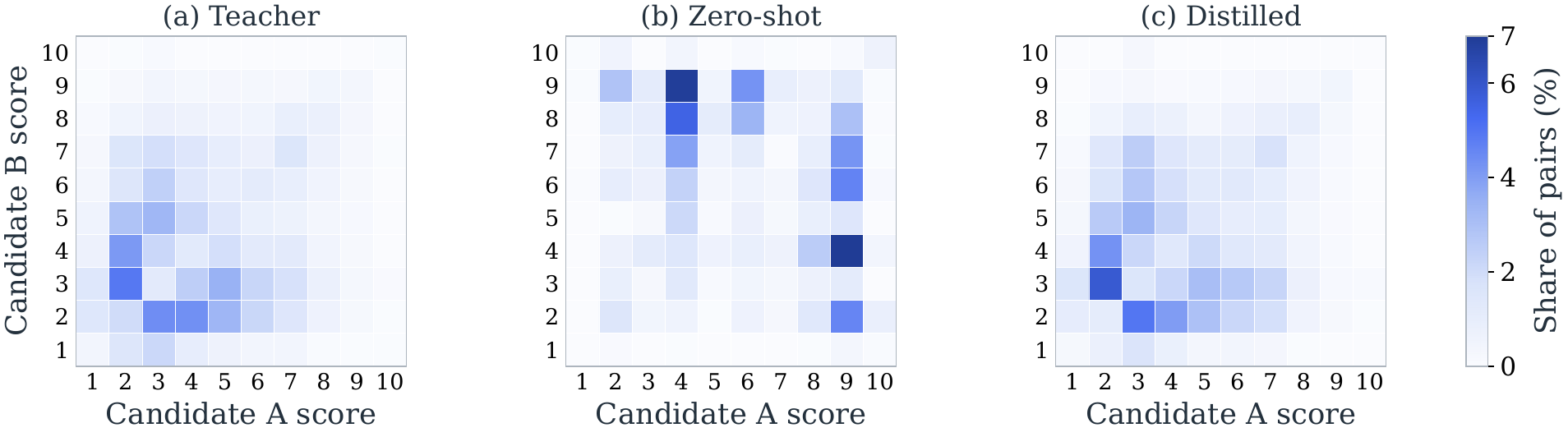}
\caption{\textbf{Distillation aligns comparative scores with the teacher.}
Joint A/B score distributions on 10,197 common valid pairs. All panels use the same percentage color scale.}
\label{fig:score-joint}
\end{figure}

\paragraph{Score statistics.}
Each pair contributes two candidate scores and one absolute score gap.
Paired score MAE averages the absolute teacher-student difference for each candidate on its matched pair.
Repeated candidate appearances across pairs are retained, giving 20,394 score observations and 10,197 gaps.
\autoref{fig:score-marginals} shows the discrete marginal proportions.
The joint heatmap in \autoref{fig:score-joint} preserves candidate A/B orientation and uses a shared linear percentage scale, with each panel summing to 100\%.
Its cells describe the two scores assigned by one model, rather than a teacher-versus-student confusion matrix.
These integer scores are comparative ratings, not calibrated task-success probabilities.

\paragraph{Common states and episode turns.}
For verification diagnostics in~\autoref{sec:judgment-analysis}, we retain states whose every comparison is valid for both models.
This leaves 1,355 of 1,624 states, containing 8,702 comparisons.
The remaining 269 states (16.6\%) are excluded because at least one stored comparison is invalid for either model.
Verification agreement is therefore conditional on complete valid responses at the state level.
States are grouped into turn bins 1--4, 5--8, 9--16, 17--32, and 33+.
The bins contain 182, 178, 388, 442, and 165 states, respectively.
The first four bins cover 21 tasks each, while the final bin covers eight.
All valid states within a bin are pooled, without task-macro weighting.

\paragraph{Computing agreement with the teacher.}
We compute both agreement metrics from saved verifier and teacher responses, without executing candidate actions.
Pairwise agreement is the fraction of comparisons where their A/B/TIE preferences match.
For verification agreement, we count each candidate's wins in the saved ring comparisons at each state, separately for the verifier and teacher.
An A/B preference gives the preferred candidate one win, while TIE gives neither candidate a win.
We identify the candidate with the most wins for each model, breaking ties by choosing the candidate that appears first in the original candidate order.
Verification agreement is the fraction of retained states where the verifier and teacher identify the same candidate.
This offline calculation uses win counts rather than the margin-weighted scores used during online execution (\autoref{sec:verification-protocols}).

\subsection{Remaining Verifier Disagreements}
\label{app:residual}

\autoref{fig:verifier-diagnostics}(b) in the main text summarizes these analyses.

\paragraph{Failure categories and reference clarity.}
Candidate semantics concerns the effect of a command or code change, and execution feasibility concerns whether it can execute as intended in the current environment.
Visible evidence concerns use of available observations, action phase concerns the timing or role of an action, and redundancy concerns repeated work.
Requirement and premature-completion failures concern unresolved task conditions and unsupported completion.
We review teacher disagreements from the zero-shot and distilled verifiers using GPT-5.6 Terra~\citep{GPT5.6}.
The judge first assesses whether the teacher preference is unambiguous and well supported by the available evidence.
Only disagreements meeting this criterion are treated as clear verifier failures for this analysis and assigned one primary category from the predefined taxonomy above.
\autoref{fig:verifier-diagnostics}(b) reports 3,328 zero-shot and 1,810 distilled cases passing this review, not the total numbers of teacher disagreements.
Candidate semantics and execution feasibility account for 1,926 zero-shot cases and 1,220 distilled cases, the latter comprising 67.4\% of the reviewed distilled-verifier failures.
These are model-judged failures relative to a teacher reference, not independently established action errors or observed trajectory failures.

\subsubsection{Examples of the Two Largest Failure Categories}
\label{app:taxonomy-examples}

These two distilled TMAX-9B disagreements illustrate the largest categories in \autoref{fig:verifier-diagnostics}(b).
Each shows an offline pairwise preference in its original A/B presentation, with GPT-5.6 Sol as reference and GPT-5.6 Terra as annotator, rather than an observed trajectory failure.

\paragraph{Candidate semantics: valid character arithmetic rejected as defective code.}
The task requires recovering a six-character token from an OCR-derived four-character prefix and two lowercase suffix characters, then deploying a C authentication server.
At turn 18, OCR suggests \texttt{WaNa}, but earlier searches have not recovered the token.
Candidate A tries the generic prefixes \texttt{BASE} and \texttt{base}.
Candidate B uses C to search lowercase suffixes and expands the prefix search to \texttt{W} followed by three alphabetic characters.
Its suffix construction includes:
\begin{lstlisting}[style=rawexample]
token[4] = 'a' + s1;
token[5] = 'a' + s2;
\end{lstlisting}
Here \texttt{s1} and \texttt{s2} range from 0 to 25.
The distilled verifier prefers A with scores 4 versus 3, claiming that \texttt{'a'+s1} is defective relative to \texttt{s1+'a'}.
These expressions are equivalent integer additions in C and generate the intended lowercase characters on the benchmark platform.
The reference instead prefers B with scores 2 versus 5 for A and B, valuing its OCR-grounded search while noting that the expanded search may exceed the command timeout.
The semantic error is the rejection of valid character arithmetic as a reason to prefer A.

\paragraph{Execution feasibility: a persistent launch rewarded despite a self-termination hazard.}
The task requires implementing a Go acoustic simulation solver, passing its regression test, and leaving an HTTP service listening on port 9090.
At turn 30, the regression test passes, but the visible state does not establish a running listener.
Candidate A builds and launches a server binary.
Candidate B runs a \texttt{python3 -c} command that scans process command lines and includes the following cleanup logic before a planned \texttt{nohup} launch:
\begin{lstlisting}[style=rawexample]
with open(f'/proc/{pid}/cmdline', 'r') as f:
    cmdline = f.read()
    if '9090' in cmdline or 'main.go' in cmdline:
        print(f'Killing PID {pid}: {cmdline[:100]}')
        os.kill(int(pid), 9)
\end{lstlisting}
The embedded Python command itself contains both matching strings, and the scan excludes neither its own process nor the invoking shell.
It can therefore terminate itself or its parent before reaching the server launch.
The distilled verifier acknowledges broad process matching but prefers B with scores 5 versus 7, rewarding its port check and persistent launch.
The reference prefers A with scores 9 versus 2, despite a separate flaw in A's port-detection logic, because B introduces this self-termination hazard.
The feasibility error is to reward the intended launch without accounting for whether execution can reach it.

\subsection{Qualitative Examples of Action Verification}
\label{app:qualitative-examples}

We present two critical steps from TMAX-27B zero-shot Mid-Harness runs.
Both use eight candidates and pairwise verification in a four-pivot tournament.
We select one recovery and one verifier error to illustrate how command-effect judgments influence the executed trajectory.

\paragraph{Recovery: a useful preference despite an imperfect rationale.}
In the React/TypeScript task, the verifier selects the one candidate among eight that directly implements the effective repair.
After the chosen action, all three Jest suites and 16 tests pass, and the external evaluator passes 17/17 tests.
Across three runs, the base agent succeeds in 0/3 and zero-shot Mid-Harness in 3/3.
The preference is useful even though the verifier incorrectly calls \texttt{diagnostics.exclude} invalid: it is a file-path filter, whereas the chosen \texttt{ignoreCodes} option filters diagnostic codes.

\paragraph{Failure: prioritizing an edit over resolving conflicting evidence.}
In the alias-canonicalization task, the verifier favors a solver rewrite over investigating a conflict between the specification and its worked example.
The chosen heuristic excludes header names from canonicalization and ultimately fails the corresponding test, leaving 11/12 tests passed and reward 0.
The traces below provide the candidate commands and original verifier responses for both cases.

\subsubsection{Detailed Action-Verification Traces}
\label{app:qualitative-examples-raw}

We expand the two examples with the task, state, candidate actions, pairwise verifier responses, and subsequent outcomes.
Long commands are shortened only where marked, while verifier reasoning is reproduced without paraphrasing.
Candidate indices refer to the original eight sampled actions, and A/B labels are local to each comparison.

\subsubsection*{Recovery Example: React and TypeScript Production Build}

\begin{tcolorbox}[colback=takeawayfill,colframe=takeawayblue,boxrule=0.6pt,arc=1pt,
  left=6pt,right=6pt,top=5pt,bottom=5pt]
\textbf{Task.}
Repair a React 18 and TypeScript 5 application so typecheck, production build, and Jest tests pass. Preserve strict mode and do not modify test files.

\smallskip
\textbf{State before verification.}
At step 28, Jest reports TS6137 for a test importing \texttt{@types/user}. The run receives reward 1 after the chosen repair.
\end{tcolorbox}

\paragraph{Candidate actions.}

\textbf{Candidate 0: separate configuration, not chosen.}
The command creates a new configuration but does not connect it to Jest.
\begin{lstlisting}[style=rawexample]
$ cat > /app/tsconfig.test.json << 'EOF'
\end{lstlisting}

\textbf{Candidate 2: active configuration with the wrong diagnostic option.}
\begin{lstlisting}[style=rawexample]
$ cat > /app/jest.config.js << 'JESTEOF'
[... unchanged mappings/config omitted ...]
      diagnostics: {
        exclude: ['TS6137'],
      },
[... closing configuration omitted ...]
JESTEOF
\end{lstlisting}

\textbf{Candidate 3: chosen.}
\begin{lstlisting}[style=rawexample]
$ cat > /app/jest.config.js << 'JESTEOF'
[... unchanged mappings/config omitted ...]
      diagnostics: {
        ignoreCodes: [6137],
      },
[... closing configuration omitted ...]
JESTEOF
\end{lstlisting}

\paragraph{Verifier reasoning: Candidate 2 versus Candidate 3.}
The verifier assigns scores 3 and 9 and prefers Candidate 3.
\begin{lstlisting}[style=rawexample,basicstyle=\rmfamily\footnotesize,backgroundcolor=\color{white},rulecolor=\color{takeawayblue!55}]
1) Requirement: The test suite fails with TS6137 because `@types/user` is interpreted as a type declaration namespace. The agent cannot modify test files, so the fix must be in the Jest/ts-jest configuration to suppress this specific error. 2) Evidence: The terminal output shows `error TS6137: Cannot import type declaration files` in `UserList.test.tsx`. 3) Candidate A attempts to use `diagnostics: { exclude: ['TS6137'] }`. This is not a valid ts-jest configuration property; `exclude` is not recognized for filtering diagnostic codes, so the error will likely persist. 4) Candidate B uses `diagnostics: { ignoreCodes: [6137] }`. This is the correct, documented ts-jest API for suppressing specific TypeScript error codes during test execution. 5) Contrast: Candidate B uses the correct configuration key (`ignoreCodes`) to resolve the blocker, whereas Candidate A uses an invalid key (`exclude`) that will fail to suppress the error, leaving the tests failing.
\end{lstlisting}

\paragraph{Verifier reasoning: Candidate 3 versus Candidate 0.}
The verifier assigns scores 9 and 2 and prefers Candidate 3.
\begin{lstlisting}[style=rawexample,basicstyle=\rmfamily\footnotesize,backgroundcolor=\color{white},rulecolor=\color{takeawayblue!55}]
The unresolved requirement is passing `npm test` despite the TS6137 error in `UserList.test.tsx` caused by importing from `@types/user`. The error message explicitly suggests importing 'user' instead, but the constraint forbids modifying test files. Candidate A updates `jest.config.js` to include `diagnostics: { ignoreCodes: [6137] }` in the ts-jest transform options. This directly suppresses the specific error code blocking the test suite, allowing the tests to run and pass. Candidate B creates a `tsconfig.test.json` but does not update `jest.config.js` to reference it (via `globals.ts-jest.tsconfig` or similar). Without updating the Jest config to use this new tsconfig, the error persists, resulting in no progress. Candidate A provides a concrete, effective fix for the immediate blocker.
\end{lstlisting}

\paragraph{Outcome.}\mbox{}
\begin{lstlisting}[style=rawexample]
Before verification:
Test Suites: 1 failed, 2 passed, 3 total
Tests:       13 passed, 13 total

After verification:
Test Suites: 3 passed, 3 total
Tests:       16 passed, 16 total

External evaluator:
17 passed in 22.77s
\end{lstlisting}

\subsubsection*{Verifier Error: Build-System Alias Canonicalization}

\begin{tcolorbox}[colback=red!2,colframe=claimred,boxrule=0.6pt,arc=1pt,
  left=6pt,right=6pt,top=5pt,bottom=5pt]
\textbf{Task.}
Implement \texttt{/app/solution.py} with \texttt{solve(lines)}, including transitive symmetric aliases and lexicographically smallest canonical names. The prompt's worked example conflicts with its explicit lexicographic rule.

\smallskip
\textbf{State before verification.}
At step 4, the implementation produces \texttt{build} for \{\texttt{build}, \texttt{compile}, \texttt{link}\}, while self-tests copied from the example expect \texttt{compile}. The run receives reward 0 after the chosen edit.
\end{tcolorbox}

\paragraph{Candidate actions.}

\textbf{Candidate 0: diagnostic inspection, not chosen.}
\begin{lstlisting}[style=rawexample]
# Alias group: {build, compile, link}
# Canonical (lex min): build (b < c < l)
#
# But expected output uses "compile" not "build"
# This means either:
# 1. The example is wrong
# 2. The canonical is NOT lex-min of the full group
print("build < compile:", 'build' < 'compile')
print("compile < link:", 'compile' < 'link')
print("build < link:", 'build' < 'link')
# build IS the lex minimum, but expected output shows compile
\end{lstlisting}

\textbf{Candidate 5: diagnostic tracing, not chosen.}
\begin{lstlisting}[style=rawexample]
# Debug test 1
import sys
from collections import defaultdict

lines = ["TARGET build", "Alias = compile link",
         "Depends = all", "TARGET compile", "Before = clean"]

# Parse
blocks = []
current_block = None
[... remaining parser/debug code omitted ...]
\end{lstlisting}

\textbf{Candidate 7: chosen.}
\begin{lstlisting}[style=rawexample]
$ cat > /app/solution.py << 'PYEOF'
[... parser and union-find omitted ...]
    canonical_map = {}
    for root, members in groups.items():
        # Among members, find those that are alias values
        value_members = [m for m in members
                         if m in alias_value_names]
        if value_members:
            canon = min(value_members)
        else:
            canon = min(members)
        for m in members:
            canonical_map[m] = canon
[... remaining implementation and self-tests omitted ...]
PYEOF
\end{lstlisting}

\paragraph{Verifier reasoning: Candidate 7 versus Candidate 5.}
The verifier assigns scores 8 and 3 and prefers Candidate 7.
\begin{lstlisting}[style=rawexample,basicstyle=\rmfamily\footnotesize,backgroundcolor=\color{white},rulecolor=\color{claimred!55}]
Requirement: The agent must fix the failing tests (alias canonicalization and topological ordering) to complete the task. Evidence: Tests 1 and 2 failed in the previous step; Test 1 shows 'build' instead of 'compile' in the output, indicating a canonicalization bug. Candidate A rewrites the entire solution with a specific fix for alias canonicalization (using lex-min of alias values) and encounter-order tie-breaking, then runs the tests. This is a concrete fix that directly addresses the observed failures. Candidate B only runs a debug script to trace the alias logic for Test 1. While debugging can be useful, the agent has already identified the likely cause (canonicalization logic) and Candidate A implements the fix immediately. Candidate B's debugging is redundant given the clear nature of the bug (lex-min of all members vs. specific subset) and delays progress. Candidate A's approach shrinks the search space by attempting a fix and verifying it, whereas B only produces more observations. A's risk is that the new canonicalization logic might be incorrect, but it is a well-reasoned hypothesis based on the spec. B provides no progress toward the final artifact.
\end{lstlisting}

\paragraph{Verifier reasoning: Candidate 7 versus Candidate 0.}
The verifier assigns scores 6 and 2 and prefers Candidate 7.
\begin{lstlisting}[style=rawexample,basicstyle=\rmfamily\footnotesize,backgroundcolor=\color{white},rulecolor=\color{claimred!55}]
Requirement: The agent must fix the failing tests (alias canonicalization and topological sort order) by implementing the correct logic, not just analyzing the problem. Evidence: Tests 1 and 2 fail in the current state; Test 1 expects 'compile' as canonical (lex-min of group {build, compile, link} is 'build', so the spec or example implies a different rule, likely that the header target is excluded from the alias group's lex-min calculation or the example implies the alias values define the group). Candidate A writes a complete new solution attempting to fix these issues by excluding header names from the canonical lex-min calculation and using encounter order for tie-breaking. This is a concrete step toward resolution, though the logic for canonicalization (excluding header names) is a heuristic guess that may be incorrect. Candidate B only runs a Python script to print comparison results and speculate about the spec being wrong. It produces no code changes to /app/solution.py. Candidate A makes progress by attempting a fix; Candidate B makes zero progress toward the goal of a working solution. A is superior because it changes the state of the artifact, whereas B is purely observational.
\end{lstlisting}

\paragraph{Outcome.}\mbox{}
\begin{lstlisting}[style=rawexample]
test_03_alias_canonicalization_lexmin

Expected:
{"cycles":[],"order":["all","build","clean"]}

Obtained:
{"cycles":[],"order":["all","compile","clean"]}

11/12 tests pass; final reward = 0.
\end{lstlisting}

The evaluator failure matches the alias rule introduced by Candidate 7.

\end{document}